\documentclass[%
 aip,
 amsmath,amssymb,
reprint, onecolumn 
]{revtex4-1}

\usepackage[T1]{fontenc}
\usepackage{mathptmx}
\usepackage{etoolbox}

\usepackage{amssymb} 
\usepackage{bm}

\usepackage{graphicx}
\usepackage{dcolumn}
\usepackage{multirow}
\usepackage{array} 
\usepackage{tabularx}
\usepackage{caption}
\usepackage{subcaption}
\usepackage{threeparttable}
\usepackage{booktabs}
\usepackage{adjustbox}

\usepackage{tikz}
\usepackage{pgfplots}
\usetikzlibrary{external, arrows.meta}
\usepgfplotslibrary{fillbetween}
\pgfplotsset{compat=1.9}
\pgfplotsset{
    every axis/.append style={
        line width=0.6pt,
    },
    mygridstyle/.style={
        grid style=solid,
        line width=0.35pt,
        color=gray!50,
    },
}

\usepackage{hyperref}
\usepackage{xcolor}
\usepackage{cleveref}

\usepackage[normalem]{ulem}
\newcolumntype{L}{>{\raggedright\arraybackslash}X} 
\newcolumntype{C}{>{\centering\arraybackslash}X}   
\definecolor{hcorange}{RGB}{245,130,48}
\definecolor{hcnavy}{RGB}{0,0,128}
\definecolor{hcblue}{RGB}{0,130,200}
\definecolor{hcpink}{RGB}{250,190,190}
\definecolor{hcbrown}{RGB}{128,0,0}
\definecolor{hclavender}{RGB}{230,190,255}
\definecolor{hcgrey}{RGB}{128,128,128}
\definecolor{hcgreen}{RGB}{60,180,75}
\definecolor{hcred}{RGB}{230, 25, 75}
\definecolor{darkred}{RGB}{128,0,0}
\definecolor{darkblue}{RGB}{0,0,170}
\definecolor{hcblack}{RGB}{0,0,0}
\definecolor{hcred2}{RGB}{255,0,0}
\definecolor{hc01}{RGB}{87,96,111}
\definecolor{hc02}{RGB}{116,125,140}
\definecolor{hc03}{RGB}{164,176,190}
\definecolor{hc04}{RGB}{255,99,109}
\definecolor{hc05}{RGB}{255,71,87}
\definecolor{hc06}{RGB}{223,228,234}
\definecolor{hc07}{RGB}{206,214,224}
\definecolor{hc088}{RGB}{241,242,246}

\definecolor{hc08}{RGB}{255,107,129}
\definecolor{hc09}{RGB}{255,120,150}
\definecolor{hc10}{RGB}{255,107,129}
\definecolor{hc11}{RGB}{255,150,150}
\definecolor{hc12}{RGB}{164,176,190}
\definecolor{hc13}{RGB}{116,125,140}
\definecolor{hc14}{RGB}{87,96,111}
\definecolor{hc15}{RGB}{47,53,66}
\definecolor{hc16}{RGB}{37,56,75}
\definecolor{hcpurple}{RGB}{229,229,255}
\definecolor{hcpurple1}{RGB}{155,110,188}
\definecolor{hc20}{RGB}{255,107,129}
\definecolor{hc21}{RGB}{164,176,190}
\definecolor{hc22}{RGB}{87,96,111}
\definecolor{hc23}{RGB}{255,71,87}
\definecolor{hc24}{RGB}{116,125,140}
\definecolor{hc25}{RGB}{47,53,66}

\crefname{equation}{Eq.}{Eqs.}
\Crefname{equation}{Eq.}{Eqs.}
\crefname{figure}{Fig.}{Figs.}
\Crefname{figure}{Fig.}{Figs.}
\crefname{algorithm}{Algorithm}{Algorithms}
\Crefname{algorithm}{Algorithm}{Algorithms}
\crefname{section}{Section}{Sections}
\Crefname{section}{Section}{Sections}
\crefname{table}{Table.}{Tables.}
\Crefname{table}{Table.}{Tables.}

\newcommand{\reply}[1]{{\color{black}{}#1}}

\makeatletter
\def\@email#1#2{%
 \endgroup
 \patchcmd{\titleblock@produce}
  {\frontmatter@RRAPformat}
  {\frontmatter@RRAPformat{\produce@RRAP{*#1\href{mailto:#2}{#2}}}\frontmatter@RRAPformat}
  {}{}
}%
\makeatother
\begin{document}

\preprint{AIP/123-QED}

\title[Optimal Parallelization Strategies for AFC in DRL-based CFD]{Optimal Parallelization Strategies for Active Flow Control in Deep Reinforcement Learning-Based Computational Fluid Dynamics}

\author{Wang Jia}
\author{Hang Xu}
\email{hangxu@sjtu.edu.cn}
\affiliation{State Key Lab of Ocean Engineering, School of Naval Architecture, Ocean and Civil Engineering, Shanghai Jiao Tong University, Shanghai, 200240, China}

\date{\today} 

\begin{abstract}
Deep Reinforcement Learning (DRL) has emerged as a promising approach for handling highly dynamic and nonlinear Active Flow Control (AFC) problems. However, the computational cost associated with training DRL models presents a significant performance bottleneck. To address this challenge and enable efficient scaling on high-performance computing architectures, this study focuses on optimizing DRL-based algorithms in parallel settings. We validate an existing state-of-the-art DRL framework used for AFC problems and discuss its efficiency bottlenecks. 
Subsequently, by deconstructing the overall framework and conducting extensive scalability benchmarks for individual components, we investigate various hybrid parallelization configurations and propose efficient parallelization strategies. 
Moreover, we refine input/output (I/O) operations in multi-environment DRL training to tackle critical overhead associated with data movement. 
Finally, we demonstrate the optimized framework for a typical AFC problem where near-linear scaling can be obtained for the overall framework. We achieve a significant boost in parallel efficiency from around 49\% to approximately 78\%, and the training process is accelerated by approximately 47 times using 60 central processing unit (CPU) cores. These findings are expected to provide valuable insights for further advancements in DRL-based AFC studies. Consequently, it continues to be a prominent and actively studied problem of significant interest.
\end{abstract}

\maketitle

\section{Introduction}

Active flow control (AFC) \citep{cattafesta2011actuators,LI202214} holds significant practical value in delaying transition \citep{liepmann1982active}, reducing separation \citep{gad1991separation}, enhancing performance \citep{akhter2021review}, lowering resistance \citep{brunton2015closed}, intensifying turbulence \citep{gad1996modern, coceal2007structure,pino2023comparative}, and suppressing noise \citep{shi2023active}.
Efficient flow control systems can improve vehicle performance and fuel efficiency while promoting economically efficient, environmentally friendly, and competitive industrial production \citep{omer2008energy, jahanmiri2010active}, thereby making flow control technology a central focus of attention in the academic and engineering communities in fields like fluid mechanics for the past several decades \citep{brunton2015closed, rizzetta1999numerical,GARNIER2021104973}.
While AFC has made significant advancements in practice, it still faces various challenges, including the design and optimization of control strategies, real-time requirements, computational complexity, uncertainties, robustness issues, and the effects of multi-physics coupling \citep{lin2016overview, collis2004issues,HE2023121376}.
Overcoming the challenges in AFC requires further research in efficient mathematical modeling, robust control strategies, and algorithm optimization to enhance system performance \citep{kral2000active, king2007active, BEWLEY200121, ClosedLoop}. 
This involves developing accurate and efficient mathematical models that capture the complex dynamics of flow control systems, designing control strategies capable of handling uncertainties and variations, and improving the computational efficiency of control algorithms \cite{811451, 1094691}.
This involves the formulation of accurate and efficient mathematical models that can effectively represent the complex dynamics of flow control systems. It also encompasses the development of control strategies capable of managing uncertainties and adapting to changes, while concurrently enhancing the computational efficiency of control algorithms \cite{811451, 1094691}. 


The rapid advancement of artificial intelligence has introduced novel approaches and implementation pathways for closed-loop AFC \cite{brunton2015closed}.
Particularly, deep reinforcement learning (DRL) methods, which facilitate autonomous learning and decision-making through interactions with the environment \cite{arulkumaran2017deep, franccois2018introduction, 9904958, francoislavet2018introduction}, align well with the requirements of closed-loop AFC \citep{gautier2015closed}.
The fundamental concept of DRL involves continuous trial-and-error learning as the intelligent agent interacts with the environment, progressively enhancing its decision-making capabilities. It utilizes deep neural networks to approximate value or policy functions, thereby mapping observations to actions. This approach, through ongoing interaction with the environment, data collection, and optimization, enables the intelligent agent to acquire the optimal decision strategy, thereby harnessing artificial intelligence algorithms to optimize and make decisions in flow control systems. In literature, \citeauthor{gazzola2014reinforcement}\cite{gazzola2014reinforcement}, \citeauthor{novati2017synchronisation}\cite{novati2017synchronisation},\citeauthor{verma2018efficient}\cite{verma2018efficient}, and \citeauthor{yan2020numerical}\cite{yan2020numerical} successively conducted a series of studies that combined numerical simulations with DRL algorithms to investigate how fish can enhance their propulsion efficiency by utilizing the wake vortices generated by other swimmers.
These promising findings provided an initial glimpse into the capability of DRL algorithms to address complex physical problems. 

Many researchers have acknowledged that computational efficiency plays a crucial role in effectively combining numerical methods with learning algorithms, such as DRL, in terms of feasibility and practicality. \citeauthor{maFluidDirectedRigid2018} are recognized as pioneers in integrating reinforcement learning with flow control\cite{maFluidDirectedRigid2018}. Their groundbreaking study on coupled control system, utilizing DRL for interactions between fluid and rigid bodies in a two-dimensional setting, has revolutionized flow control and paved the way for innovative implementations of AFC. \citeauthor{rabault2019artificial} achieved remarkable results in AFC by training an artificial neural network using a DRL agent, successfully stabilizing vortex shedding and reducing resistance by approximately 8\%\cite{rabault2019artificial}. \citeauthor{rabault2019accelerating} further addressed the limitation of training speed caused by the computational fluid dynamics (CFD) component in training DRL agents by proposing a parallelization solution that combines parallelization techniques for both the numerical simulation and the DRL algorithm, resulting in significant speedup\cite{rabault2019accelerating}. Thanks to the open-source code released by \citeauthor{rabault2019artificial} and his team\cite{rabault2019artificial}, there has been a remarkable surge in scholarly publications utilizing DRL algorithms to address fluid dynamics challenges, especially since 2020. Until now, DRL has demonstrated successful applications in various AFC tasks, encompassing the 1D Kuramoto–Sivashinsky equation \citep{xuReinforcementlearningbasedControlConvectivelyunstable2023}, 1D falling fluid flow \citep{belus2019exploiting}, 2D Rayleigh–B\'{e}nard convection \citep{beintemaControllingRayleighBenard2020}, 2D convection control in confined space \citep{wangClosedloopForcedHeat2023}, 3D model cavity with one heat source \citep{wangControlPolicyTransfer2024}, 2D flow over a cylinder oscillating around its axis \citep{tokarev2020deep} and the interaction between rotating small cylinders and the wake of a main cylinder \citep{xu2020active}.

Despite the proclaimed advantages of DRL-based methods, various challenges persist and prevent widespread implementation of these methods. One major bottleneck is the excessive training cost associated with the DRL-based framework. As reported by \citeauthor{rabault2019artificial}, their codebase allows to train the artificial neural network to perform AFC on a 2D cylinder in about 24 hours, using a modern CPU running on a single core\cite{rabault2019artificial}.
\citeauthor{doi:10.2514/6.2018-3691} utilized the DRL framework for a similar control problem where they spent 48 hours on training, using a 32-core processor\cite{doi:10.2514/6.2018-3691}. Evidently, to tackle more complex and demanding AFC problems, the cost of training will quickly become prohibitively expensive and parallelization techniques are necessary to scale the framework for larger problems. 
Unfortunately, the scaling performance of DRL-based CFD problems remains a largely unexplored research field. \citeauthor{rabault2019accelerating} discovered that the DRL agents can learn from multiple environments launched in parallel with similar rate of convergence, but without any scaling analysis, it remains unclear whether multi-environment training translates into a proportional reduction in the actual training time\cite{rabault2019accelerating}.
The closest performance analysis on this topic in literature is the work by \citeauthor{kurzDeepReinforcementLearning2022}, who introduced an Reinforcement Learning-augmented CFD solver, showing 30\% parallel efficiency on 16 nodes in a HPC system\cite{kurzDeepReinforcementLearning2022}.
Nonetheless, to the best of our knowledge, there is no existing work that comprehensively analyzes the optimal form of parallelization for maximum efficiency in a DRL-based AFC problem.  
In this work, we want to provide a thorough analysis for different forms of parallelization, which is supported by scaling performance benchmarks, in order to identify performance bottlenecks and find ways to further optimize the current state of the art. Ultimately, through tackling performance issues, we wish to shed some light on the best practices for optimal DRL-based training in the context of AFC problems. 

The outline of this paper is as follows.
In \cref{sec:Methodology}, the problem description is presented, followed by an in-depth overview of DRL and its formulation in addressing the AFC challenge. The available parallelization strategies in DRL-based AFC problems are also described, along with a discussion of their software implementations. In \cref{sec:results}, a validation study is conducted, followed by a detailed scaling analysis of both CFD simulations and the DRL framework within multi-environment settings. This analysis is then extended to the examine hybrid parallelization techniques that strive to balance CFD parallelization and DRL parallelization for optimal efficiency. In particular, the potential for further performance enhancement through the refinement of \reply{input/output (}I/O\reply{)} operations in \cref{sec:Further optimization through refining I/O operations} is also investigated. Finally, \cref{sec:Conclusions} concludes the paper by summarizing the key findings, discussing their implications, and suggesting potential directions for future research.

\section{Problem description and methodology}\label{sec:Methodology} 

Here, we investigate the application of a DRL-based AFC technique to control flow separation over a circular cylinder, including the numerical setup, DRL algorithm, formulation of the AFC problem, hybrid parallelization strategy, and the software platform integrating CFD and DRL.

\subsection{Numerical simulation}\label{sec:numerical simulation}

The configuration adopted in this work for simulating the flow around a circular cylinder closely follows the classical benchmark by \citeauthor{Schafer1996}\cite{Schafer1996}
We consider the two-dimensional, unsteady, incompressible and viscous flow around a circular cylinder. The corresponding governing equations are the Navier-Stokes equations (in their dimensionless form)
\begin{equation}\label{eq:ns1}
\frac{\partial \boldsymbol{u}}{\partial t}+\boldsymbol{u} \cdot(\nabla \boldsymbol{u})=-\nabla p+\frac{1}{Re} \Delta \boldsymbol{u},
\end{equation}
\begin{equation}\label{eq:ns2}
\nabla \cdot \boldsymbol{u}=0.
\end{equation}
Here, $\boldsymbol{u} = (U, V)$ represents the fluid velocity, where $U$ and $V$ are the velocity components along the $x$- and $y$-axes, respectively. $t$ denotes the dimensionless time unit. $p$ is the thermodynamic pressure. $Re = \frac{\overline{U}D}{\nu}$ is the Reynolds number, where $\overline{U}$ is the mean velocity at the inlet, $D$ is the diameter of the cylinder and $\nu$ is the kinematic viscosity. 

\begin{figure}[ht]
\centering
\begin{subfigure}{\textwidth}
\centering
\includegraphics[width=0.95\linewidth]{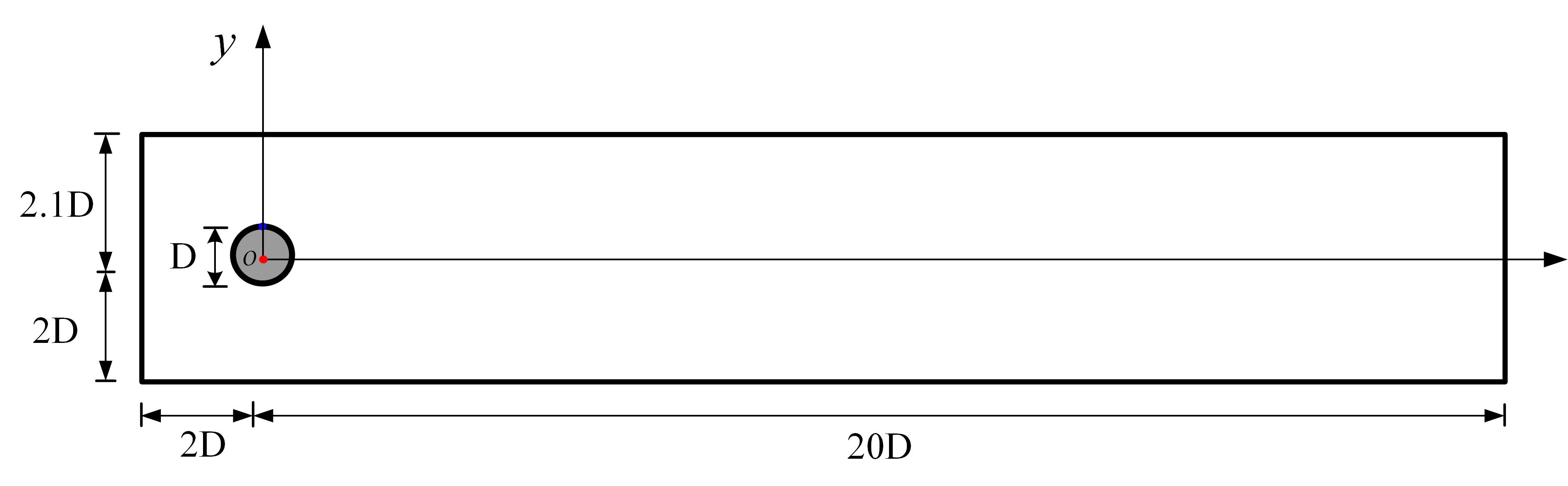}
\caption{}
\label{fig1:cfd01}
\end{subfigure}
\begin{subfigure}{0.5\textwidth}
\centering
\includegraphics[width=\linewidth]{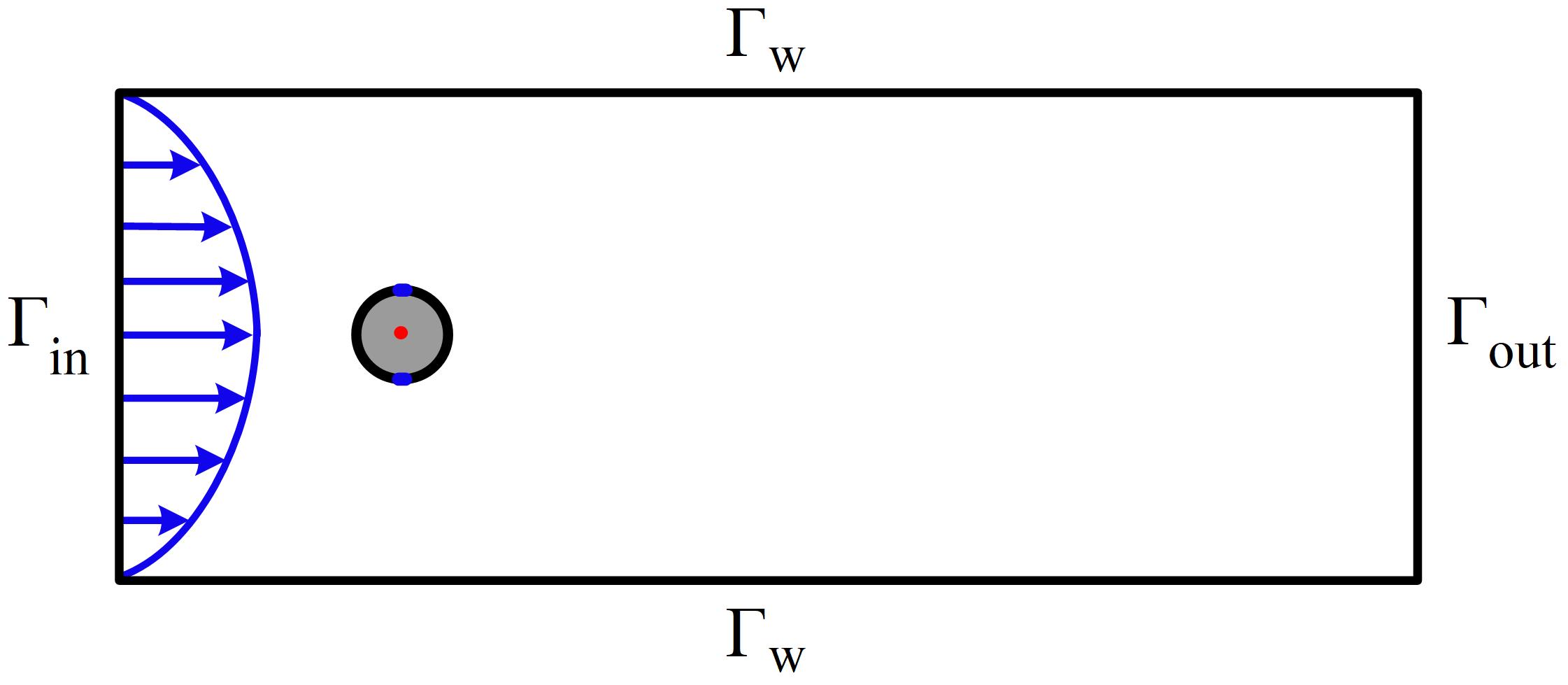}
\caption{}
\label{fig1:cfd02}
\end{subfigure}
\qquad\qquad 
\begin{subfigure}{0.26\textwidth}
\centering
\includegraphics[width=\linewidth]{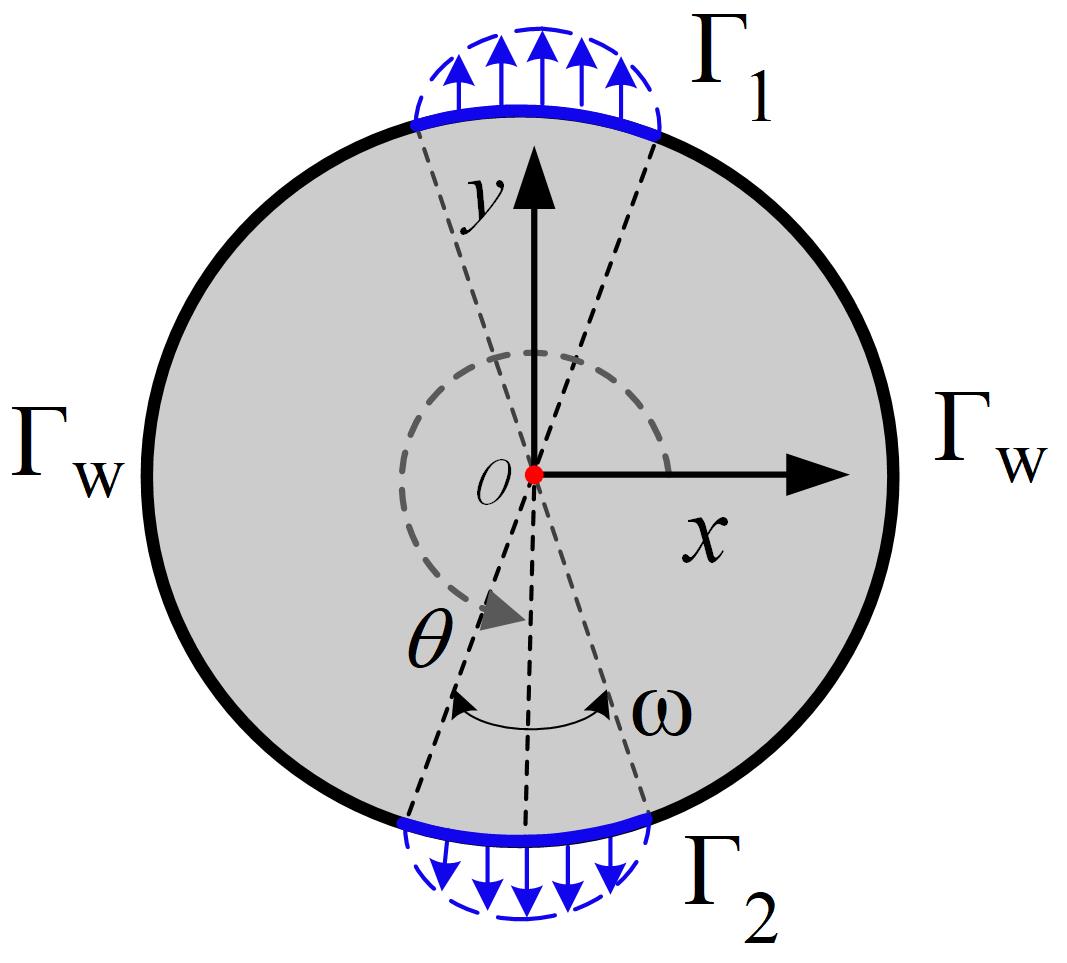}
\caption{}
\label{fig1:cfd03}
\end{subfigure}
\caption{ Description of the numerical setup. (a) Computational domain ($22D \times 4.1D$) of flow around a cylinder. (b) Boundary conditions for the computational domain; (c) Details of the boundary conditions on the cylinder. Jets are located at $\Omega = 90^{\circ}$ and $ \Omega = 270^{\circ}$ on the cylinder, with a jet width of $\omega = 10^{\circ}$. Parabolic velocity distribution is used for each jet.} 
\label{fig:cfd}
\end{figure}

As shown in \cref{fig1:cfd01}, the cylinder is submerged in a rectangular domain with dimensions of $22D$ (along the $x$-axis) and $4.1D$ (along the $y$-axis), with the origin of the Cartesian coordinate system established at the center of the cylinder. The cylinder is positioned slightly off-centered in the y-direction in order to trigger vortex shedding. 
The boundaries of the computational domain are divided into an inlet $\Gamma_\text{in}$, an outlet $\Gamma_\text{out}$, no-slip walls $\Gamma_\text{w}$, and two separate jets on the cylinder $\Gamma_{i}\, (i = 1,2)$, as shown in~\cref{fig1:cfd02}. 
At the inlet $\Gamma_\text{in}$, the inflow velocity along $x$-axis is prescribed by a parabolic velocity profile in the form,
\begin{equation}\label{equ:1velocity}
U_{\text{inlet}}(y)=U_m\frac{(H-2y)(H + 2y)}{H^2},
\end{equation}
and that along $y$-axis is prescribed as,
\begin{equation}\label{equ:2velocity}
V_{\text{inlet}}(y) = 0.
\end{equation}
$U_m$ is the maximum velocity magnitude of the parabolic profile, and $H=4.1D$ represents the total height of the rectangular domain. The average inlet velocity $\overline{U}$, is related to the parabolic velocity profile $U_{\text{inlet}}(y)$ through the expression:
\begin{equation}\label{equ:mean velocity}
\overline{U}=\frac{1}{H} \int_{-H/2}^{H/2} U_{\text{inlet}}(y) d y=\frac{2}{3} U_m.
\end{equation}

At the outlet  $\Gamma_\text{out}$, an outflow boundary condition is applied where the velocity is to be extrapolated. No-slip condition is applied at both the upper and lower walls $\Gamma_\text{w}$.
On the surface of the cylinder, we implement AFC by introducing synthetic jets at the highest and lowest points of the cylinder, as shown in~\cref{fig1:cfd03}. 
These jets $\Gamma_{i}\, (i = 1,2)$ can be used to inject or suction fluid for flow control. 
Each jet has a width of $\omega = 10^{\circ}$ and is directly perpendicular to the outer surface of the cylindrical wall. 
The velocity distribution within each jet follows a parabolic profile.
The jet velocities can be positive or negative, corresponding to blowing or suction, respectively.
It is crucial to maintain a balance in the net mass flow rates of all the jets to ensure that the net flow into or out of the system is zero, i.e., $V_{\Gamma_1} = -V_{\Gamma_2}$. This constraint guarantees overall conservation of mass within the system. No-slip boundary condition is applied on the surface of the cylinder where the jets are not located.

The aforementioned problem is solved using the open-source CFD software \texttt{OpenFOAM}, written in \texttt{C/C++}\cite{jasakOpenFOAMLibraryComplex2013}.
Specifically, the built-in solver, PimpleFoam which combines the Pressure-Implicit with Splitting of Operators (PISO) algorithm and the Semi-Implicit Method for Pressure-Linked Equations (SIMPLE) algorithm, is employed to solve this incompressible fluid flow problem\cite{ISSA198640,patankar1980numerical,SIMPLE}.
The computational domain is discretized into 16,200 grid elements using unstructured triangular mesh units. 
The mean velocity magnitude is selected as $\overline{U}$= 1, and the Reynolds number is set to $Re$ = 100. To ensure numerical stability, the time step is chosen as $\Delta t$ = 0.0005. The simulation results \reply{for this numerical setup} were verified against the classical benchmark \reply{in a previous study by \citeauthor{wangDRLinFluids}\cite{wangDRLinFluids}.} 
In this work, we are interested in comparing the lift coefficient ($C_L$) and drag coefficient ($C_D$), which are defined as
\begin{align}
     C_L = \frac{F_L}{0.5\rho \overline{U}^2D}, \quad  C_D = \frac{F_D}{0.5\rho \overline{U}^2D}.
\end{align}
Here, $F_L$ and $F_D$ represent the lift and drag forces integrated on the surface of the cylinder, respectively, and $\rho$ is the fluid density.

\subsection{Overview of DRL}\label{sec:DRL control}
\begin{figure}[ht]
\centering
\includegraphics[width=0.8\textwidth]{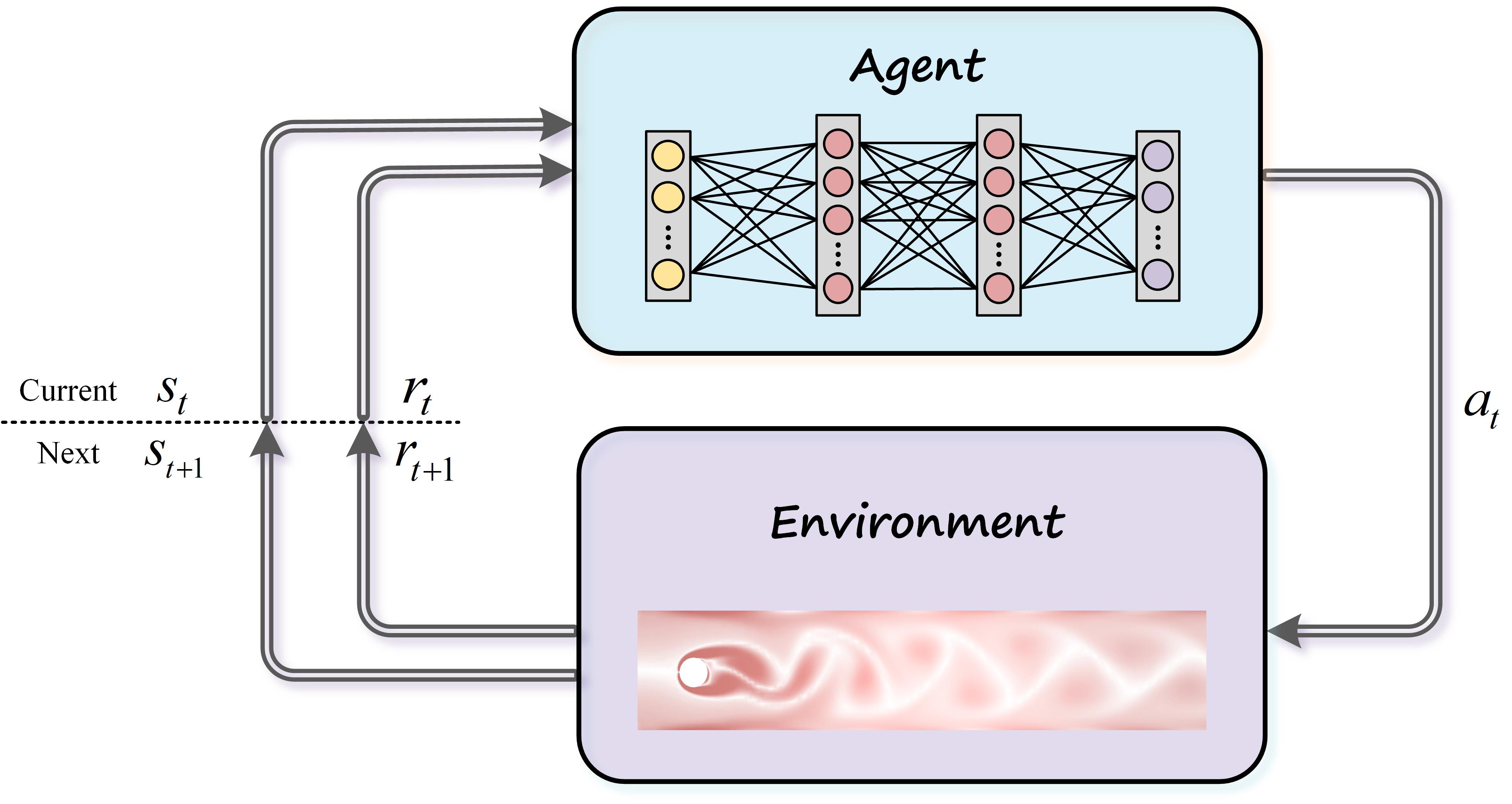}
\caption{The agent and the environment are fundamental components of reinforcement learning. The environment is the entity with which the agent interacts. At any given time step $t$, the agent first observes the current state $s_t$ of the environment, along with the corresponding reward value $r_t$. Based on these state and reward information, the agent decides how to take action $a_t$. The agent receives feedback from the environment, obtaining the next time step's state $s_{t+1}$ and reward $r_{t+1}$.}
\label{fig:RL}
\end{figure}

This subsection provides a brief overview of DRL. However, this is by no means an exhaustive summary and readers are referred to ~\citeauthor{arulkumaranBriefSurveyDeep2017,ladoszExplorationDeepReinforcement2022,shakyaReinforcementLearningAlgorithms2023}~\cite{arulkumaranBriefSurveyDeep2017,ladoszExplorationDeepReinforcement2022,shakyaReinforcementLearningAlgorithms2023} for a more thorough discussion. 
The reinforcement learning problem is typically modeled as a Markov decision process (MDP), which is a mathematical framework that formalizes sequential decision-making problems involving an agent interacting with an environment\cite{bellman1957markovian}. \cref{fig:RL} illustrates the typical interaction between an agent and an environment.
At any time-step $t$, the environment is at some state $s_t$. Based on this observing state, the agent's policy $\pi_\theta(a_t | s_t)$ determines the agent's action $a_t$, which in turn causes the environment to transit into a new state $s_{t+1}$\cite{puterman1990markov,bellman1957markovian}.
At the same time, a reward $r_t$ is awarded to the agent. The ultimate goal is to find the optimal policy that maximizes the agent's cumulative reward over the long run. 
By actively collecting and learning from this data, the agent can enhance its performance over time by improving its policy optimization and decision-making\cite{van2012reinforcement,arulkumaran2017deep,8103164}.

The Proximal Policy Optimization (PPO) algorithm \cite{schulman2017proximal,pmlr-v37-schulman15,heess2017emergence,schulman2017trust} , which is a specific implementation based on the Policy Gradient approach for optimizing an agent's policy function in reinforcement learning, is used for training in this study.
The PPO algorithm enhances training stability by introducing a \textit{proximal optimization} mechanism \cite{OPT-003,9904958,arulkumaran2017deep} and effectively addresses challenges in continuous action spaces and high-dimensional state spaces, leveraging the power of deep learning for efficient processing and enabling deep decision-making models to excel in complex tasks\cite{9904958,arulkumaran2017deep}.

For a given trajectory $\tau=\left\{\left(s_0, a_0\right),\left(s_1, a_1, r_1\right), \ldots \ldots,\left(s_n, a_n, r_n\right)\right\}$ that describes a sequence of states  from the initial state, through various states and actions, the end state, the expected value of the discounted return $G(\tau)$ is obtained by maximizing the following objective function:\cite{NIPS1999_464d828b}
\begin{equation}\label{eq:cumulative reward}
J\left(\theta\right)=\underset{\tau \sim \pi_\theta}{\mathbb{E}}[G(\tau)],
\end{equation}
which quantifies the expected cumulative reward that an agent can achieve by following the policy $\pi_\theta$. Here $\mathbb{E}$ denotes the expectation over the trajectory $\tau$ sampled from the policy $\pi_\theta$, and $G(\tau)$ is the discounted return function at a specific time step $t$, defined as:
\begin{equation}\label{eq:reward function}
G(t)=\sum_{k=0}^T \gamma^k R_{t+k+1},
\end{equation}
where $\gamma \in [0,1]$ is the discount factor that determines the relative importance of future rewards compared to immediate rewards, and $T$ is the total length of the game.
The optimal parameter $\theta^*$ is obtained as:\cite{NIPS1999_464d828b}
\begin{equation}\label{eq:maximize}
\theta^* = \underset{\theta}{\text{argmax}} \underset{\tau \sim \pi_\theta}{\mathbb{E}}[G(\tau)]. 
\end{equation}
This equation represents the search for $\theta$ that maximizes the expected value of the discounted return by sampling trajectories $\tau$ from the policy $\pi_\theta$.
For the purpose of iterative updates, the objective function of the PPO algorithm is typically defined as a surrogate objective that balances exploration and exploitation,\cite{schulman2017proximal} as follows:
\begin{equation}\label{eq:ppo}
\left.L^{C L I P}(\theta)=\hat{\mathbb{E}}_t\left[\min \left(r_t(\theta)\right) \hat{A}_t, \operatorname{clip}\left(r_t(\theta), 1-\varepsilon, 1+\varepsilon\right) \hat{A}_t\right)\right],
\end{equation}
where $L^{C L I P}(\theta)$ represents the objective function that is to be maximized or minimized. $\theta$ represents the parameters of the policy being optimized. 
$\hat{\mathbb{E}}_t$ represents the expectation operator, computing the average value of the clipped surrogate objective term over sampled trajectories with respect to the policy's probability distribution.  $r_t(\theta)$ is the ratio between the probabilities of selected actions under the new and old policies. $\varepsilon$ is a hyperparameter. $\hat{A}_t$ denotes the advantage estimate, which measures the disparity between the expected value and the value of a specific action in a given state. The $\operatorname{clip}(.)$ function controls the scale of updates during training by constraining the ratio between the probabilities of the new and old policies within a specified range $[1-\varepsilon, 1+\varepsilon]$.
Note that in this equation, the first term within the min function is 
employed in calculating the objective function for policy updates, and the second term in the equation, $\operatorname{clip}\left(r_t(\theta), 1-\varepsilon, 1+\varepsilon\right) \hat{A}_t$, modifies the surrogate objective by constraining the probability ratio\cite{schulman2017proximal}.

\subsection{Formulation of the AFC problem using DRL}\label{sec:Formulation of the AFC problem using DRL}

The relationship between the DRL component and the CFD component is outlined in this subsection. We follow the coupling strategy proposed by \citeauthor{wangDRLinFluids}\cite{wangDRLinFluids}. For a more comprehensive understanding, we encourage readers to consult their work and the references therein.

The overarching goal of the AFC problem is to achieve drag reduction on the cylinder by suppressing vortex shedding in the von K\'{a}rm\'{a}n vortex street. To address this problem using DRL, the CFD numerical simulation serves as the environment with which the agent interacts. Specifically, the agent interacts with the environment through three components: state, action and reward.

\begin{figure}[ht]
\centering
\includegraphics[width=0.82\linewidth]{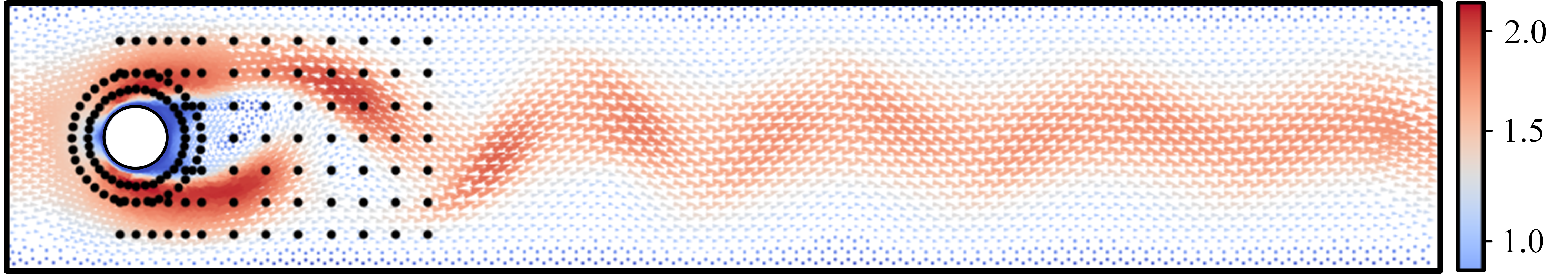}
\caption{Instantaneous velocity field with black dots representing the positions of the 149 probes.}
\label{fig:probes}
\end{figure}

\begin{itemize}
    \item At each moment in time, the state consists of the instantaneous flow field data gathered from specific positions called probes within the computational domain. A comprehensive collection of 149 probes is strategically placed around the cylinder and in the wake region\reply{, and their positions are shown in \cref{fig:probes}}
    These probes are carefully positioned to capture crucial flow characteristics while maintaining a sparse distribution. \reply{For validation purposes, the probe layout in this work is adopted from that proposed by \citeauthor{wangDRLinFluids}\cite{wangDRLinFluids} Other designs of the probe layout can also be found in literature\cite{rabault2019artificial,parisRobustFlowControl2021,liReinforcementlearning,castellanosMachine}.} 
    \item Based on the observed state, the agent's policy determines the action. The agent's policy is parameterized by a two-layer artificial neural network with 512 neurons in each layer, \reply{as advised by \citeauthor{rabault2019artificial}\cite{rabault2019artificial}}. The resulting action value, denoted as $a$, is responsible for adjusting the velocities of the synthetic jets. This enables precise control over the flow dynamics.
    To ensure stability and avoid non-physical instabilities in the flow, a smoothing function is applied, which holds the following relationship:
    \begin{equation}
        V_{\Gamma_1,T_i} = V_{\Gamma_1,T_i} + \beta\left(a - V_{\Gamma_1,T_{i-1}}\right),
    \end{equation}
    where $V_{\Gamma_1,T_i}$ represents the jet velocity at the end of the $i$th actuation control period $T_i$, and $\beta$ is a numerical parameter used for smoothing. The value of $\beta$ is determined through trial and error, and in this case, it is set to $\beta = 0.4$. Each control period $T_i$ consists of 50 time steps, denoted as $T_i = (50i)\Delta t$. To prevent excessive energy input in the jets, an upper limit constraint is imposed, ensuring that $V_{\Gamma_i} \leq U_m$.
    \item At the end of each control period, a reward function is employed to evaluate the agent's performance. With the primary aim of achieving maximum drag reduction, the reward function is defined as
    \begin{equation}\label{eq:my reward}
        r_{T_i}=C_{D,0}-\left(C_D\right)_{T_i}-\omega\left|\left(C_L\right)_{T_i}\right|,
    \end{equation}
    where $C_D$ and $C_L$ are the drag and lift coefficients, respectively. The term $C_{D,0}$ corresponds to the mean drag coefficient of the circular cylinder without any flow control, specifically set to $C_{D,0} = 3.205$. The notation $(\cdot)_{T_i}$ indicates an averaging operation over the actuation period $T_i$. The parameter $\omega = 0.1$ serves as a weighting factor that emphasizes the contributions of lift fluctuations in the reward calculation. The inclusion of the lift term is intended to prevent unnecessary increases in lift fluctuations during the optimization process.
    
\end{itemize}

In the training process, each episode consists of a complete CFD simulation that spans 5000 time steps. This duration corresponds to 100 actuation periods. Therefore, the maximum total time for each episode is $T_{\max} = T_{100} = 2.5$ non-dimensional time units.
The length of the episode is intentionally selected to encompass multiple vortex shedding cycles. This allows the agent to effectively learn the control algorithm by observing and adapting to the dynamics of the flow over several cycles.

\subsection{Parallelization of DRL-based AFC problem}\label{sec:Parallelization of DRL-based AFC problem}

In a previous study, it was demonstrated that approximately 99\% of the computational time is dedicated to CFD simulations. With increasingly complex simulations, there is an intrinsic demand for utilizing parallelization in DRL training to reduce the overall runtime\cite{rabault2019accelerating}.

\begin{figure}[ht]
    \centering
    \includegraphics[width=\textwidth]{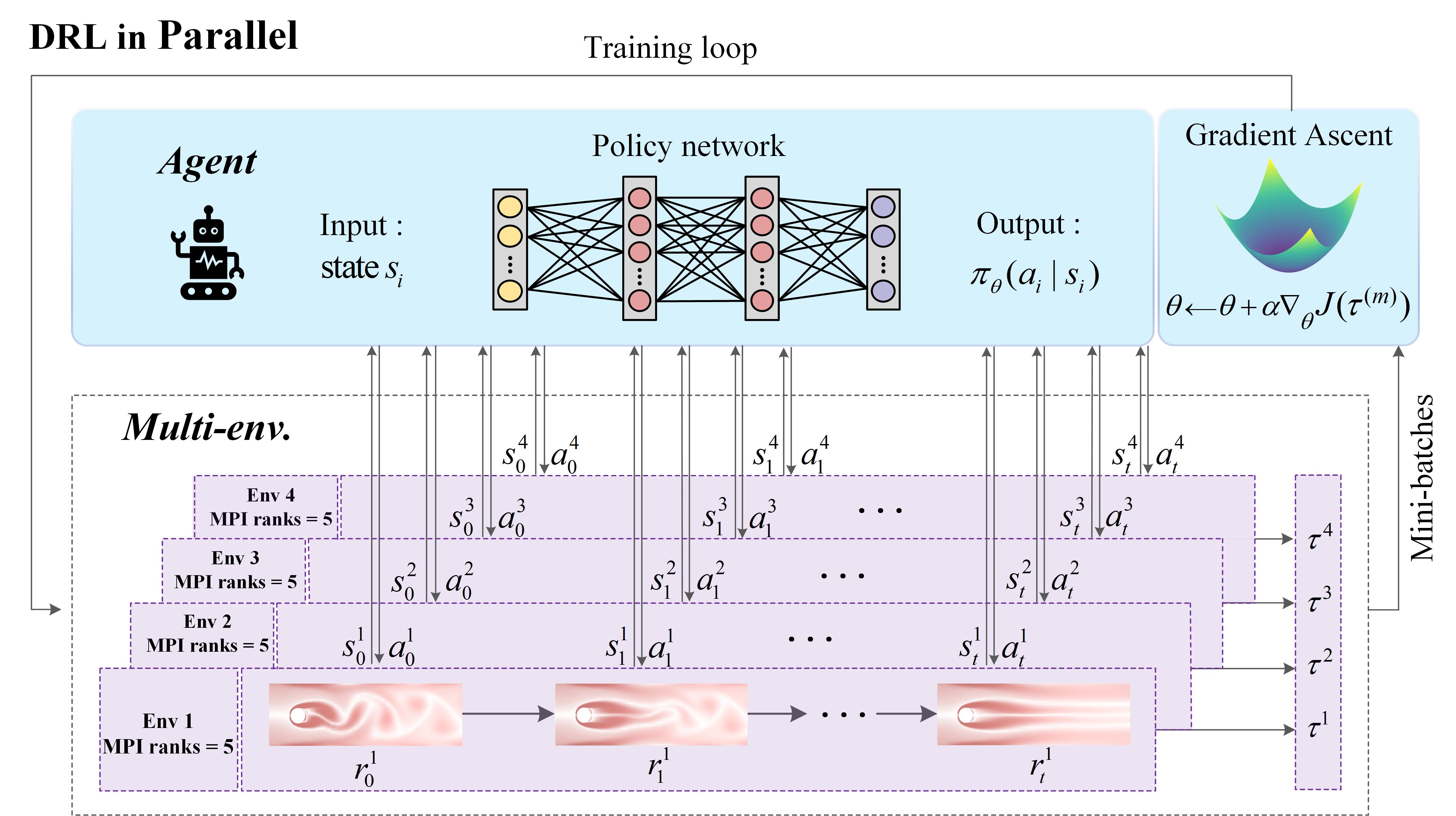}
    \caption{Illustration of the process and allocation of computational resources for DRL parallel training. Here, four training environments are utilized, each employing 5 MPI ranks for parallel CFD computations. This configuration requires a total of 20 cores. During the training, the agent continuously interacts with the CFD environments, generating a series of tuples ($s_i^m$, $a_i^m$, $r_i^m$). Here, the superscript $m$ denotes the specific environment, and the subscript $i$ represents the timestep when the agent interacts with that environment. The trajectory $\tau^m$ resulting from the agent-environment interaction is then used to calculate the gradient $\nabla_\theta J(\tau^{(m)})$ and update the neural network parameters $\theta$. DRL parallelizes the training across the four environments, meaning that in each training step, the four environment instances operate simultaneously. This allows for the concurrent collection of multiple samples. Importantly, each environment instance operates independently, creating an isolated setting for training.}
    \label{fig:agent}
\end{figure}

In a general DRL-based framework for AFC,  there are two available parallelization strategies: parallelizing the CFD simulation and parallelizing the DRL training process through multiple environments. The CFD component can allow parallelization through parallel programming softwares, such as Message Passing Inteface (MPI) where one CFD instance is distributed across several processes. On the other hand, the data collection process in DRL can be parallelized by creating multiple environments simultaneously. Each environment represents a separate instance of the AFC problem, allowing the DRL agent to interact and learn from multiple environments concurrently. 

The hybrid parallelization approach used for DRL parallel training is illustrated below. Each environment consists of multiple concurrent processes (or MPI ranks) to launch CFD simulations. Typically, the number of processes or MPI ranks ($N_\text{ranks}$) is equal to the number of CPUs ($N_\text{CPUs}$) dedicated to a single CFD simulation. This setup accelerates the simulation time for each actuation period, enabling faster transitions between states according to the agent's actions and reducing the time required to collect ($s, a, r$)-tuples along each trajectory.
Furthermore, training is conducted simultaneously in multiple environments, which expedites data collection and processing, thus speeding up the overall training process. At any given time, multiple environments with multiple CFD instances can coexist, each having different flow states. Each environment operates independently and in parallel, with states influenced solely by the history of actions within that environment. This parallelization allows for concurrent state transitions and reward calculations, providing a larger number of samples and experiences for the DRL algorithm within a given timeframe. Once all environments complete one training episode, data from multiple trajectories are batched together in mini-batches to update weights and parameters in the DRL algorithm.
In a hybrid setup with $N_\text{envs}$ environments, the total number of CPUs for the overall framework is the product of the number of ranks per CFD simulation ($N_\text{ranks}$) and the number of environments ($N_\text{envs}$), i.e., $N_\text{total CPUs} = N_\text{envs} \times N_\text{ranks}$. Figure \ref{fig:agent} presents an example of hybrid parallelization using a total of 20 CPUs, with 4 environments ($N_\text{envs} = 4$) and 5 MPI ranks ($N_\text{ranks} = 5$). The framework allows flexibility in choosing the number of CPUs and environments for either form of parallelization, without any specific restrictions.

\subsection{Software platform}\label{sec:software platform}

We build our work upon the codebase \texttt{DRLinFluids} developed by \citeauthor{wangDRLinFluids}\cite{wangDRLinFluids}. As mentioned in \cref{sec:numerical simulation}, the CFD component is implemented using the open-source software \texttt{OpenFOAM}\cite{jasakOpenFOAMLibraryComplex2013}. The DRL component is implemented using another open-source Python framework \texttt{TensorForce}\cite{tensorforce}. Both components allow for parallelization as discussed in \cref{sec:Parallelization of DRL-based AFC problem}.

To establish the communication between the CFD simulation software and the DRL package, \citeauthor{wangDRLinFluids} proposed the utilization of file \reply{I/O} streams as the interface for these two components within the framework of \texttt{DRLinFluids}\cite{wangDRLinFluids,ritchie1984unix}.
At the end of each actuation period, the observed states, histories of aerodynamics coefficients (which are necessary for reward computation), and resulting flow fields are saved from memory to the hard disk. 
Subsequently, the state and history files are parsed into Python scripts to be further processed by the DRL package. 

The action values generated by the DRL algorithm are subsequently extracted and integrated back into the \texttt{OpenFoam} configuration files using Regex (Regular expression) operations\cite{10.1145/363347.363387}. These updated configuration files, along with the most recent flow field data, are used to initiate the subsequent instance of \texttt{OpenFoam} for simulating the next actuation period. Therefore, each episode of training with 100 actuation periods consist of 100 instances of \texttt{OpenFoam} processes interleaved with I/O to/from the Python package \texttt{TensorForce}\cite{lift-tensorforce,tensorforce}.
This iterative process ensures continuous interaction and synchronization between the DRL algorithm and the CFD simulation environment. \citeauthor{wangDRLinFluids}\cite{wangDRLinFluids} argued that the adoption of an I/O-based interface is suitable, since the primary performance bottleneck lies in the computation in numerical simulations rather than the data movements in I/O.
In \cref{sec:DRL parallel efficiency}, we will evaluate the validity of this assumption rigorously, highlight its shortcomings, and propose improvements to further optimize these interfaces.

\section{Results}\label{sec:results}

The results' analysis is as follows: In~\cref{sec:DRL in AFC}, the control results achieved by \citeauthor{wangDRLinFluids}\cite{wangDRLinFluids} are initially replicated, with a particular emphasis on training efficiency and performance. Subsequently, the entire framework is deconstructed, and the parallel efficiency of each individual component is meticulously analyzed. In \cref{sec:CFD parallel efficiency}, the MPI-based parallelization of the CFD solver is delved into, while \cref{sec:DRL parallel efficiency} focuses on multi-environment parallel training within the DRL framework. To further optimize the framework and overcome bottlenecks in I/O operations, refinements are explored in \cref{sec:Further optimization through refining I/O operations}. 

Throughout the analysis, computational hardware equipped with an Intel(R) Xeon(R) Platinum 8358 CPU @ 2.60GHz, featuring a total of 64 cores (32 cores per socket across 2 sockets), is employed. The software suite utilized includes \texttt{TensorForce} version 0.6.0 as the reinforcement learning framework and \texttt{OpenFOAM} version 8 as the CFD platform.

\begin{figure}[ht]
\includegraphics[width=0.97\textwidth]{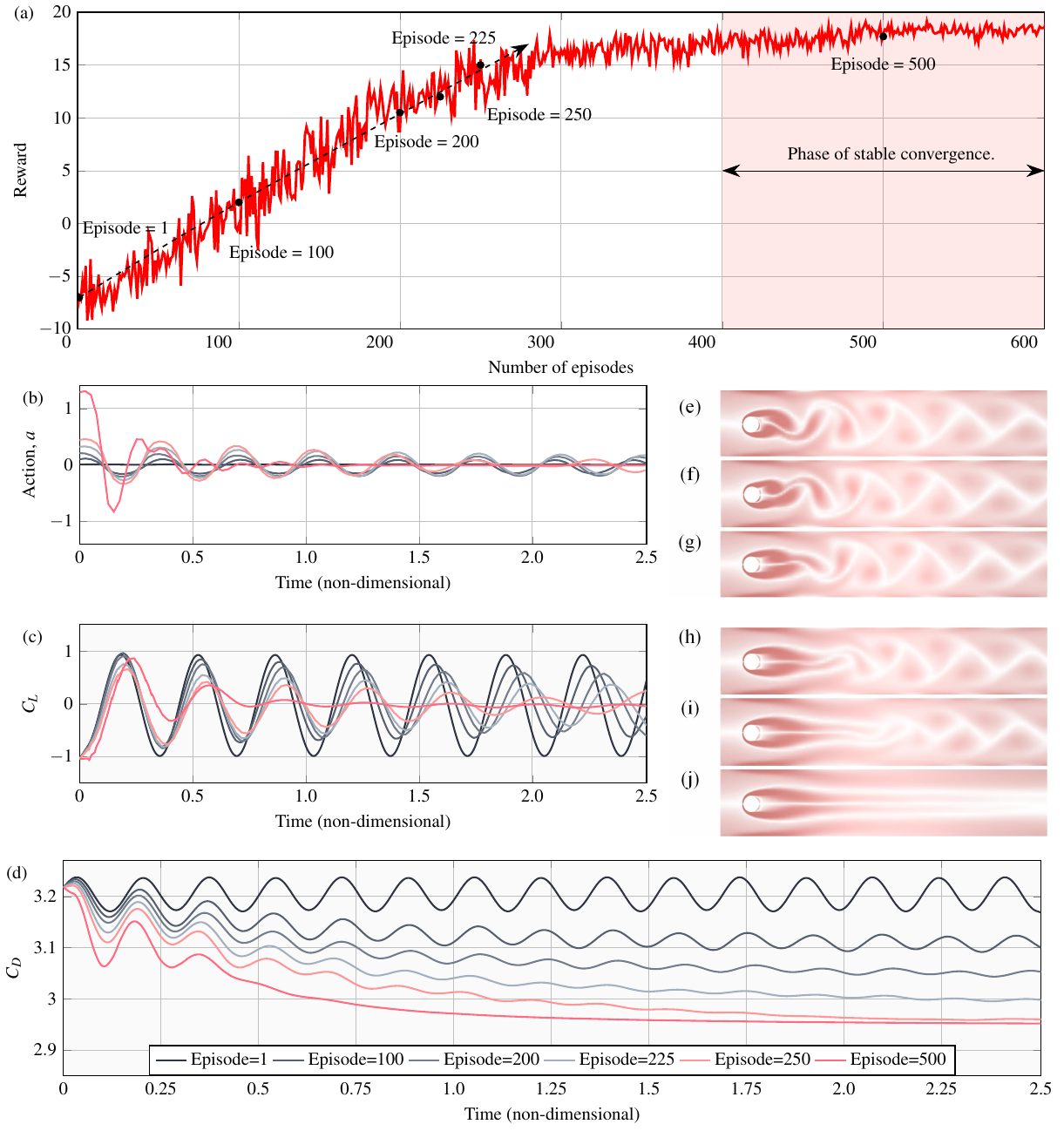}
\caption{Results of DRL training.(a) displays the cumulative reward at each episode during the training process. (b), (c), and (d) show the changes in action $a$, lift coefficient $C_L$ and drag coefficient $C_D$, respectively. Histories of each parameter at selected episodes are shown to illustrate the convergence. (e) to (j) correspond to vorticity contours at the end of each selected episode.}
\label{fig:training}
\end{figure}

\subsection{Validation study}\label{sec:DRL in AFC}

The \texttt{DRLinFluids} framework proposed by \citeauthor{wangDRLinFluids}\cite{wangDRLinFluids} is validated here. We apply the DRL algorithm to learn \reply{AFC} strategies, with a specific focus on the application of synthetic jet excitation in the presence of flow around a circular cylinder. The results are summarized in \cref{fig:training}. Our findings demonstrate the effectiveness of the DRL algorithm in reducing flow resistance around a circular cylinder through AFC. These results serve as evidence of the capability of the \texttt{DRLinFluids} framework in achieving the desired control objectives.

During training, the PPO algorithm iteratively optimizes the parametrized neural network-based policy to maximize cumulative rewards at each episode. Fig. 5(a) displays the learning curve, which depicts the development trend of the return for each round. Starting from an initial reward value, the agent's return gradually increases and tends to converge around 300 episodes. This tortuous rise in the learning curve is a direct consequence of the reinforcement learning process, where the agent continuously experiments, makes mistakes, accumulates experience, and gradually strengthens the reward function. 

To illustrate the learning process, we focus on selected episodes and analyze the action value ($a$), lift coefficient ($C_L$), and drag coefficient ($C_D$) as shown in  Fig. 5(b) to  Fig. 5(d). In the first episode, the agent starts with small action values, leading to minimal changes in the flow. Consequently, the lift and drag coefficients remain similar to those in the uncontrolled flow. As the number of episodes increases, the agent learns to explore the action space to optimize the objective. By the 100th episode, there are indications that the agent is learning to introduce flow changes periodically, aligning with the shedding frequency of vortices. This control strategy successfully reduces drag and yields increased reward.
By the 200th episode, the agent further improves the reward through better drag reduction policies. Notably, at the 225th and 250th episodes, significant optimization of drag reduction is observed within a few training episodes. This suggests that the agent capitalizes on learned knowledge from previous episodes and focuses more on exploitation than exploration to find the optimal policy.
After 400 episodes, the reward converges to a steady value. At the 500th episode, the drag and lift coefficients ($C_D$ and $C_L$) demonstrate the successful suppression of the vortex shedding phenomenon. The vorticity contours in  Fig. 5(b) to  Fig. 5(d) further illustrate the complete suppression or control of vortex shedding behind the circular cylinder.
Overall, the implementation of DRL control results in a significant decrease in the drag coefficient ($C_D$), reducing it from 3.205 to approximately 2.95. This corresponds to a remarkable drag reduction rate of 8\%. Additionally, our proposed control strategy effectively mitigates the periodic fluctuation of the lift coefficient ($C_L$).

In the context of run-time performance, our focus is on the DRL-based AFC framework. In this validation study, we utilized a single reinforcement learning environment, with CFD simulations running on a single core. The training process consisted of 600 episodes, resulting in a total training time of 45.1 hours, with each episode taking approximately 4.5 minutes. Comparatively, \citeauthor{wangDRLinFluids}\cite{wangDRLinFluids} also employed a single environment training for the same AFC problem. They reported an average time of 10 minutes per episode on a 22-core CPU, resulting in a total training time of 100 hours for 600 episodes. While the total time consumption may vary due to differences in computer hardware, it is evident that both studies require a significant amount of time to achieve stable convergence results. 
Based on the previous work \citeauthor{rabault2019artificial}\cite{rabault2019artificial} and our experiences (as mentioned in \cref{sec:Further optimization through refining I/O operations}), it is observed that the computational time of CFD simulations constitutes the majority (over 95\%) of the total time. For complex flow phenomena, the computational demands of CFD simulations can become prohibitively expensive for single-core computation. Therefore, leveraging parallel computing architectures is crucial to optimize the run-time performance and make DRL feasible for complex AFC problems.

\begin{figure}[htbp]
\includegraphics[width=\textwidth]{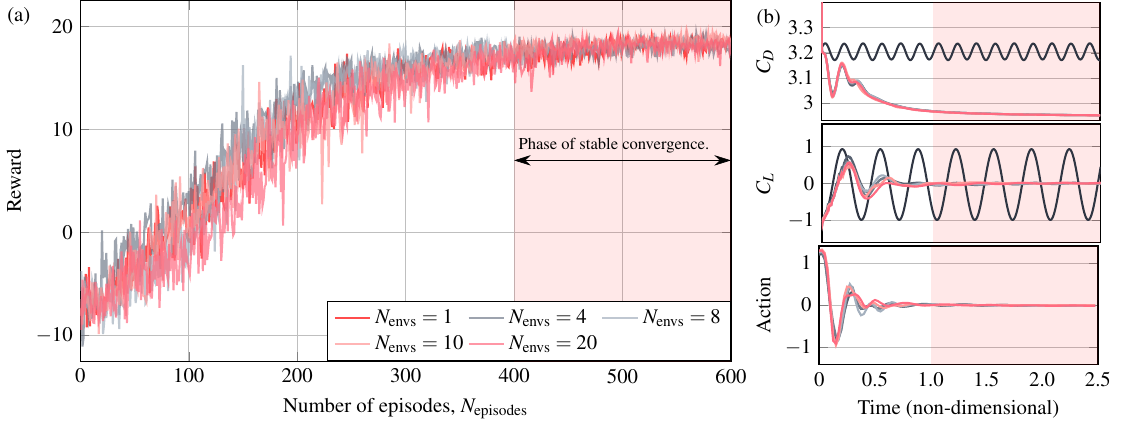}
\caption{(a) shows the reward function curve when the number of parallel environments $N_\text{envs}$ = 1, 4, 8, 10, 20. (b) shows the final results of training when different numbers of environments are run in parallel. Drag coefficient $C_D$, lift coefficient $C_L$, and action values are consistent with the single-environment training in \cref{fig:training}.}
\label{fig:multi_envs}
\end{figure}

In a previous study\cite{rabault2019accelerating}, parallelization through multi-environment training has demonstrated potential in expediting the training duration. Given that the convergence of rewards appears to be largely unaffected by the quantity of environments, initiating numerous training environments could feasibly diminish the time required to accumulate learning samples essential for the agent's training. In \cref{fig:multi_envs}, we verify this claim using our framework and observe that the convergence rate of rewards is indeed consistent across various scenarios. Drag and lift coefficients are controlled successfully in all cases. Under the assumption that training duration for each episode remains the same, this could imply a reduction in total training time proportional to the increase in the number of environments. However, it is crucial to acknowledge that the duration of each training episode may not remain consistent in a parallel setting due to communication overhead and additional complexities inherent in multi-environment training. Furthermore, the prospect of parallelizing each training episode through the application of multi-core CFD simulations has not been thoroughly explored in existing literature.

Therefore, in a parallel setting, the question of how best to balance the parallelization of CFD simulations with the use of multiple DRL environments for optimal performance is still open to exploration. To the best of our knowledge, no existing work provides a comprehensive analysis for the scaling performance of DRL frameworks when applied to AFC problems.
In the subsequent subsections, we shall analyze various parallelization strategies, computing resource allocation in different scenarios, and other bottlenecks that hinder performance. Our objective is to identify strategies that effectively utilize computing resources, minimize training time, and provide valuable insights for optimizing the design of parallel training and computation in AFC applications.

\subsection{Scaling analysis of CFD simulations}\label{sec:CFD parallel efficiency} 

In this subsection, we shift our attention to the scaling efficacy of CFD computations with the objective of identifying the most suitable core allocation for parallelized simulations.
In an effort to isolate the CFD component from potential DRL-related disturbances, this set of numerical experiments are conducted where only a single environment is used for training, i.e., $N_\text{total CPUs} = N_\text{ranks}$. It should be noted that the parallel efficiency of \texttt{OpenFoam},\cite{starikoviciusEfficiencyAnalysisOpenFOAMBased2016a,towaraMPIParallelDiscreteAdjoint2015,culpo2011current,janesMPIAssociatedScalability2022,liuParallelPartitionedApproach2021} along with CFD algorithms at large,\cite{capuanoComparativeStudySpectralelement2019,WITHERDEN20143028,CANTWELL2015205} has been extensively studied in literature. Despite this, to date, there lacks an investigation specifically targeted at CFD parallelization within the framework of DRL applications. For instance, given that DRL can be parallelized through multi-environment training, no clear guidelines exist on how to balance the hybrid parallelization between CFD computations and DRL training. 
Moreover, integrating CFD solvers within a DRL framework introduces distinct characteristics, such as frequent interactions with the DRL module, which could theoretically impact the scaling performance. 

Taking all these factors into consideration, we benchmark the parallel efficiency of CFD computations to establish a foundation for discussing the overall scalability of the framework. To achieve this, we employ a multi-core computing approach and make adjustments to the core count or number of number of MPI ranks, $N_\text{ranks}$, for CFD computations. This allows us to evaluate the performance of the CFD solver under different parallelization configurations and provide insights into the scalability of the framework.
We systematically record the computation time for each $N_\text{rank}$ and calculate the relationship between speedup and the number of ranks. This allows us to gain insights into the effectiveness of parallel computing of the CFD component and determine the number of $N_\text{rank}$ ranks that yields optimal performance. For investigating the effect of interaction with DRL, we present two sets of scaling results. The first set with $T_1 = 0.025$ denotes scaling performance of a single instance of \texttt{OpenFoam}. This evaluates the parallelism for the CFD solver by itself. The other set with $T_\text{max}=T_{100} = 2.5$ denotes scaling performance where we consider the entire training episode with 100 instances of \texttt{OpenFoam} and intermediate data movement with the DRL component. This evaluates the parallel performance of the CFD solver in a realistic setting when coupled with DRL frameworks.

\begin{figure*}[ht]
    \centering
    \begin{subfigure}{0.49\textwidth}
    \includegraphics[width=\textwidth]{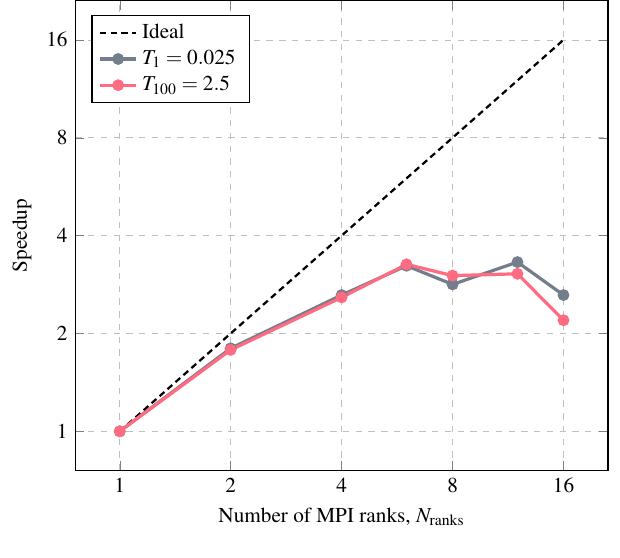}
    \caption{}
    \label{fig:actual_hours1}
    \end{subfigure}
    \begin{subfigure}{0.49\textwidth}
    \includegraphics[width=\textwidth]{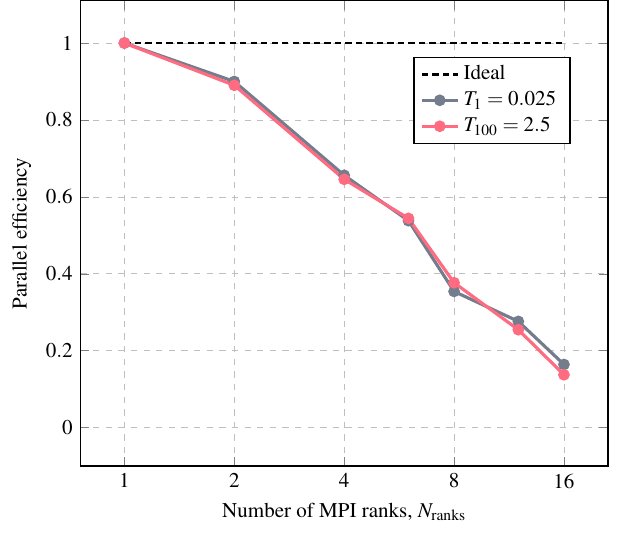}
    \caption{}
    \label{fig:actual_hours2}
    \end{subfigure}
\caption{Scaling performance of the CFD solver: (a) speedup ratio, (b) parallel efficiency of the CFD instances. $N_\text{ranks}=1$ in each set of data is taken as the respective reference point. $T_1$ denotes one instance of the CFD solver. $T_{100}$ denotes  100 instances of the CFD solver interleaved with data transfer with the DRL framework, which is equivalent to one episode of single-environment training.}
\label{fig4:CFDparallelefficiency}
\end{figure*}

In \cref{fig4:CFDparallelefficiency}, we present the influence of the number of MPI ranks ($N_\text{ranks}$) on the parallel speedup ratio and efficiency for CFD computations. In an ideal scenario, the speedup ratio would exhibit a linear relationship with $N_\text{ranks}$, where the speedup ratio equals the number of cores employed. However, as shown in \cref{fig:actual_hours1}, the current CFD parallel computations demonstrate a lower speed ratio compared to the ideal value. Even when two $N_\text{ranks}$ are utilized in parallel, the speedup drops significantly and fails to reach 2. Further analysis reveals that as $N_\text{ranks}$ increases, the incremental gain diminishes and eventually saturates.
In addition, \cref{fig:actual_hours2} demonstrates that the parallel efficiency decreases rapidly as the number of MPI ranks increases. With a 2-core parallel scheme, the utilization of computational resources can only reach 90\%. As $N_\text{ranks}$ further increases, the parallel efficiency drops below 20\% when $N_\text{ranks} = 16$. This drastic decrease in parallel efficiency indicates that allocating resources to parallelize individual \texttt{OpenFOAM} instances has limited performance benefits. It is worth noting that both sets of benchmarks show similar performance, suggesting that the interaction with the DRL component does not introduce significant overhead under single-environment training. The discussion regarding multi-environment training will be addressed in \cref{sec:Further optimization through refining I/O operations}.

We acknowledge that it is possible to optimize the parallel performance of the CFD solver through algorithmic and software development. However, such optimization is highly dependent on various factors, including the numerical scheme and mesh topology, and it remains an active research field. Therefore, the performance optimization of the CFD solver for the application of DRL to generic AFC problems is considered beyond the scope of the current work. The results presented above are characteristic and provide a sufficient foundation for the upcoming discussions.

\subsection{Scaling analysis of the DRL framework with multi-environments}\label{sec:DRL parallel efficiency} 

Following the analysis of the CFD component in the previous section, this segment focuses on the parallel efficiency of the DRL component. Our objective is to comprehensively evaluate the impact of varying the number of environments on the performance of parallel training and explore the relationship between the number of environments in DRL training and the speedup ratio. To achieve this, we employ three different scenarios for the parallelization of CFD, utilizing one, two, and five MPI ranks, respectively. This enables us to investigate the influence of different computing resources on the parallel efficiency of the DRL framework.

\subsubsection{Multi-environment DRL parallelization}

In the first set of tests, we allocate five MPI ranks for each environment to conduct the flow simulations using CFD. This configuration is to analyze the parallel efficiency of DRL when CFD parallelization is not optimal. All tests are executed on a 64-core CPU, where we concurrently utilize up to 12 environments to parallelize the DRL component. (Four cores are not used in this case.) For each test, encompassing a training volume of $N_\text{episodes}=3000$ episodes, we document the duration required for the training process and evaluate its parallel efficiency. Results of these tests are consolidated and presented in \cref{tab:125cores_sumup}. As observed in \cref{tab:125cores_sumup}, increasing the number of environments from 1 to 12 while keeping the number of MPI ranks constant at 5 per environment results in a decrease in training duration from 305.8 hours to 32.4 hours. This trend is indicative of successful parallel scaling up to a certain point. However, the parallel efficiency exhibits a slight decrement from 100\% down to 78.6\% as the number of environments increases. 

\begin{table}[ht]
\centering
\caption {Statistics of parallel multi-environment training with $N_\text{ranks}=1,2,5$.}
\vspace{-\baselineskip}
\begin{tabularx}{\textwidth}{CCCCCCc}
\toprule
$N_\text{episodes}$ & $N_\text{envs}$ & $N_\text{ranks}$ & $N_\text{total CPUs}$ & Total duration (h) & Speedup & Parallel efficiency (\%) \\
\midrule
\multicolumn{7}{c}{\textbf{CFD MPI Ranks, $N_\text{ranks}=5$ }} \\
\midrule
3000 & 1 & 5 & 5 & 305.8 & 1.0 & 100.0 \\
3000 & 2 & 5 & 10 & 170.8 & 1.8 & 89.5 \\
3000 & 4 & 5 & 20 & 88.5 & 3.5 & 86.4 \\
3000 & 6 & 5 & 30 & 59.7 & 5.1 & 85.3 \\
3000 & 8 & 5 & 40 & 47.3 & 6.5 & 80.8 \\
3000 & 10 & 5 & 50 & 38.3 & 8.0 & 79.8 \\
3000 & 12 & 5 & 60 & 32.4 & 9.4 & 78.6 \\
\midrule
\multicolumn{7}{c}{\textbf{CFD MPI Ranks, $N_\text{ranks}=2$}} \\
\midrule
3000 & 1 & 2 & 2 & 289.6 & 1.0 & 100.0 \\
3000 & 2 & 2 & 4 & 156.3 & 1.9 & 92.6 \\
3000 & 4 & 2 & 8 & 80.0 & 3.6 & 90.4 \\
3000 & 6 & 2 & 12 & 53.4 & 5.4 & 90.3 \\
3000 & 8 & 2 & 16 & 40.8 & 7.1 & 88.7 \\
3000 & 10 & 2 & 20 & 33.2 & 8.7 & 87.1 \\
3000 & 20 & 2 & 40 & 17.7 & 16.4 & 81.9 \\
3000 & 30 & 2 & 60 & 12.4 & 23.3 & 77.5 \\
\midrule
\multicolumn{7}{c}{\textbf{CFD MPI Ranks, $N_\text{ranks}=1$}} \\
\midrule
3000 & 1 & 1 & 1 & 225.2 & 1.0 & 100.0 \\
3000 & 2 & 1 & 2 & 123.7 & 1.8 & 91.1 \\
3000 & 4 & 1 & 4 & 64.6 & 3.5 & 87.1 \\
3000 & 6 & 1 & 6 & 44.4 & 5.1 & 84.5 \\
3000 & 8 & 1 & 8 & 33.9 & 6.6 & 83.0 \\
3000 & 10 & 1 & 10 & 26.3 & 8.6 & 85.5 \\
3000 & 20 & 1 & 20 & 14.2 & 15.9 & 79.3 \\
3000 & 30 & 1 & 30 & 9.6 & 23.5 & 78.4 \\
3000 & 40 & 1 & 40 & 9.0 & 25.0 & 62.4 \\
3000 & 50 & 1 & 50 & 8.1 & 27.7 & 55.4 \\
3000 & 60 & 1 & 60 & 7.6 & 29.6 & 49.3 \\
\bottomrule
\end{tabularx}
\label{tab:125cores_sumup}
\end{table}

It is noteworthy that the parallel efficiency remains around 80\% even as the number of environments increases to 12, which suggests that the DRL component scales relatively well with the addition of computational resources. 
The notable efficiency in the multi-environment setup is largely due to the independent execution of CFD simulations across different environments, which run concurrently without data dependency. Interactions between the CFD simulations and the DRL framework within each trajectory incur minimal communication costs since extracting policy actions from the DRL framework is computationally inexpensive compared to CFD simulations. Post-trajectory updates in the DRL framework, while slightly more communication-intensive, are still less demanding than the CFD simulations themselves. Additional factors such as the management of multi-threading processes and file I/O operations may contribute marginally to the overall computational overhead. Nevertheless, The dominant CFD simulations are almost embarrassingly parallel with respect to the number of environments, leading to remarkable scaling performance. 

We emphasize that the scaling analyses detailed above utilized a multi-core configuration for CFD with $N_\text{ranks}=5$. As outlined in \cref{sec:CFD parallel efficiency}, such a configuration does not represent an optimal allocation of computational resources, attributed to the limited parallel efficiency of the CFD software. Nonetheless, we observe good scaling performance for the DRL component as we increase the number of environments. 
In \cref{tab:125cores_sumup}, we present two further sets of experimental outcomes wherein $N_\text{ranks}$ are configured to 1 and 2, respectively. The speedup ratio for all tests are visualized in~\cref{fig:125_sumup}. As explained in \cref{sec:CFD parallel efficiency}, simulations conducted on a single-core with $N_\text{ranks}=1$ are the most efficient, whereas dual-core simulations with $N_\text{ranks}=2$ are sub-optimal. Across all configurations, we observe a consistently robust scaling performance. The scaling performance for $N_\text{ranks}=1$ drops slightly for very large-scale runs. We investigate this performance regression in \cref{sec:Further optimization through refining I/O operations}. Nevertheless, the parallel efficiency is significantly better than those presented in \cref{sec:CFD parallel efficiency}. The implications of these findings are twofold: firstly, the parallel efficiency of multi-environment training appears to be invariant to the parallelization of the underlying CFD simulations; secondly, prioritizing the parallelization of the DRL component over that of CFD simulations is a more resource-efficient strategy.

\subsubsection{Resource allocation for hybrid parallelization}

Here, we evaluate resource allocation strategies with the intent to ascertain the most efficient hybrid parallel configuration.
The objective is to provide a road-map for leveraging hybrid parallelism in DRL-infused CFD simulations.
This requires a reassessment of the configurations involving one, two, and five MPI ranks for CFD parallelization, in conjunction with a variable number of environments, to analyze their scaling performance against the total number of cores employed. A standardized reference point is essential for a consistent comparison on resource utilization efficiency. For this purpose, we utilize the test case with a single-core CFD and single environment ($N_\text{rank}=1, N_\text{envs}=1$) as the reference baseline for calculating the parallel speedup and efficiency across the entire framework. 

\begin{figure}[htbp]
    \centering
    \includegraphics[width=0.6\textwidth]{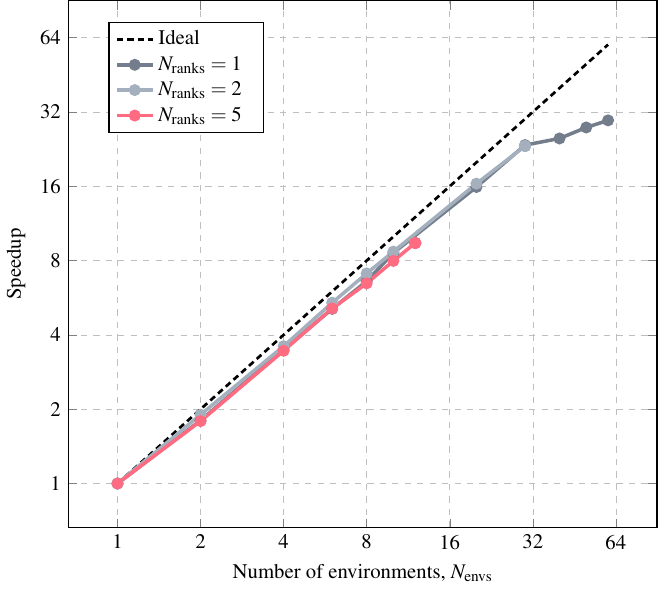}
    \caption{Speedup ratio of the multi-environment DRL training. Three different configurations for the CFD solver are used: $N_\text{ranks}=1,2,5$. $N_\text{envs}=1$ in each set of data is taken as the respective reference point.}
    \label{fig:125_sumup}
\end{figure}

\begin{figure*}[ht]
    \centering
    \begin{subfigure}{0.49\textwidth}
    \includegraphics[width=\textwidth]{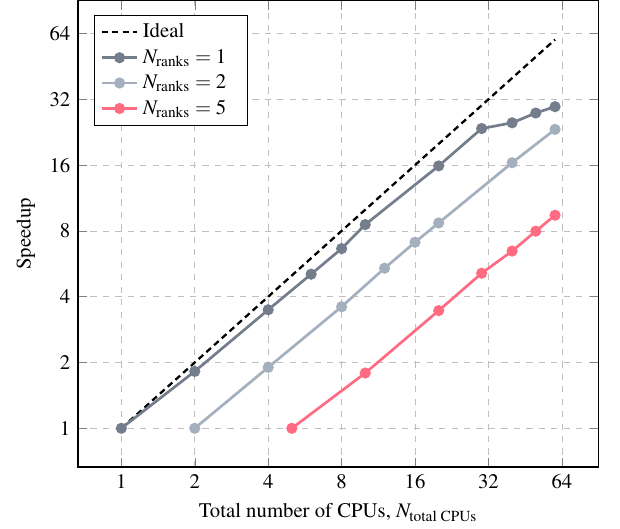}
    \caption{}
    \label{fig7:speedup}
    \end{subfigure}
    \begin{subfigure}{0.49\textwidth}
    \includegraphics[width=\textwidth]{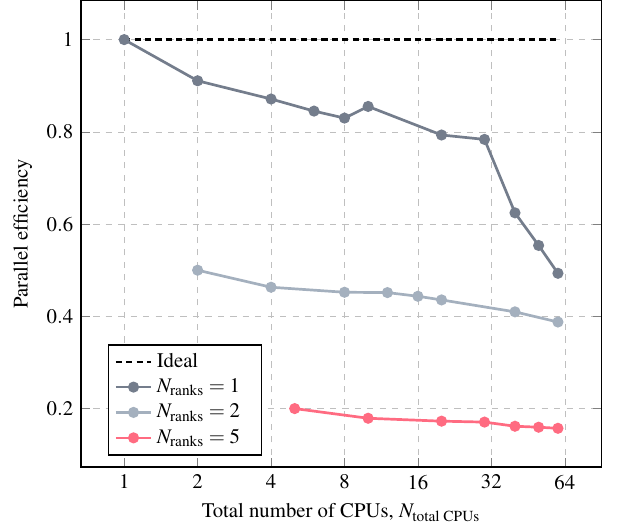}
    \caption{}
    \label{fig7:parallel}
    \end{subfigure}
  \caption{Scaling performance of the multi-environment DRL training: (a) speedup ratio (b) parallel efficiency. Three different configurations for the CFD solver are used: $N_\text{ranks}=1,2,5$. \{$N_\text{ranks=1}, N_\text{envs}=1$\} is taken as the reference point \textit{for all data points}.}
  \label{fig7:efficiency}
\end{figure*}

The corresponding results are presented in \cref{fig7:efficiency}. 
Specifically, in the speedup plot in \cref{fig7:speedup}, the speedup ratios increase as the number of CPU cores increases. The gradients of each curve illustrate the efficiency of parallelizing DRL through multiple environments and agree with those in \cref{fig:125_sumup}. The consistency in the gradient provides proof that multi-environment parallelization is largely unaffected by the underlying CFD parallelization configuration due to the embarrassingly parallel nature of these environments.
The offsets among different curves indicate the efficiency of CFD parallelization. With the same total number of CPUs, using multiple ranks for CFD lowers the speedup ratio as the CFD component with \texttt{OpenFoam} has poor parallel performance. For a more direct comparison, \cref{fig7:parallel} details the total parallel efficiency for all cases. When CFD utilizes five-rank MPI parallel computation, the parallel efficiency remains consistently below 20\%. When CFD adopts two-rank MPI parallel computation, the parallel efficiency decreases from 50\% to around 38\%. However, when CFD relies on a single-core computation, the parallel efficiency declines from 100\% to approximately 49\%. Despite these declining trends, single-core configuration still exhibits superior parallel efficiency compared to other parallelization schemes.

The empirical evidence suggests that the most optimal parallelization strategy involves DRL multi-environment parallel training coupled with CFD single-core computation.
Our experiments utilizing 60 cores demonstrate that the duration for training can be reduced from 225.2 hours to a mere 7.6 hours. This equates to a speedup of approximately 30-fold, substantiating the effectiveness of this hybrid parallelization approach.

\subsection{Further optimization through refining I/O operations}\label{sec:Further optimization through refining I/O operations}

\begin{figure}[htbp]
    \centering
    \includegraphics[width=\textwidth]{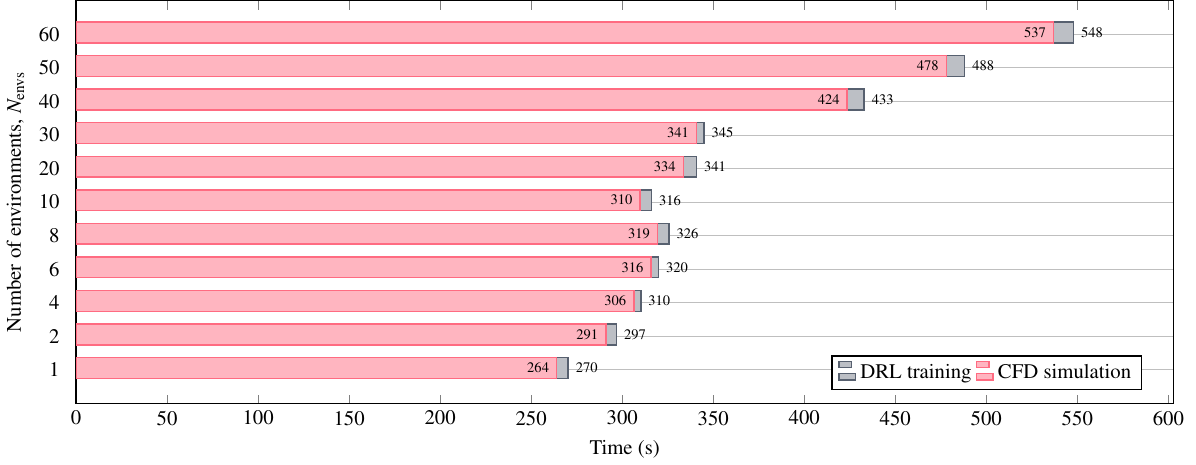}
    \caption{Time breakdown for one episode of training. CFD simulation time predominates in all cases, but increases rapidly after $N_\text{envs} > 30$. The DRL training cost, which consists of policy update and interfacing with the CFD solver, is minimal compared to CFD simulation time.}
\label{fig:barfig}
\end{figure}

In the preceding subsection, we notice that when CFD employs single-core computation, utilizing multi-environment parallel training in DRL achieves the optimal parallel acceleration. However, as the number of parallel environments in DRL is substantially increased ($N_\text{total CPUs} > 30$), there is a drastic decrease in parallel efficiency. Such a decrease is not evident when employing $N_\text{ranks}=2$ or $5$. This pattern implies that the regression in performance may be linked to the expansive number of environments. 
To further explore this issue, we conducted a profiling of the computational time required for a single training episode, with the results shown in \cref{fig:barfig}. The data indicates that the CFD simulation time predominates the training duration. However, there is an unexpected and significant rise in the time taken for CFD simulations when the number of environments is very large, even though these times should remain fairly consistent due to the absence of data dependencies among the environments. We postulate that this downturn may be attributed to the substantial involvement of I/O operations associated with multi-environment parallel training. 

To verify this hypothesis, we design three sets of experiments, namely \textit{Baseline}, \textit{I/O-Disabled}, and \textit{Optimized}, for comparative analysis. All experiments use single-core CFD computation.
Results obtained in the previous subsection are used as the baseline (\textit{Baseline}), where the training time reduces from 225.2 hours for a single environment to 7.6 hours for 60 parallel environments. \reply{It is worth mentioning that multiple files with a total size of 5.0 MB are generated at the end of each instance of CFD simulation within each environment. For 60 parallel environments with 100 actuation periods within an episode, this is equivalent to 300 MB of CFD-related I/O at the end of each actuation period and at least 30 GB of data transfer within each episode.}

In the second experimental set (\textit{I/O-Disabled}), we determine the theoretical upper bound of performance by completely disabling all I/O operations within the workflow. Although this configuration deviates from a realistic operational setting -- foregoing the essential data exchange between distinct CFD simulations and the DRL framework -- it provides an exploratory benchmark indicative of the maximum attainable performance in the absence of I/O overhead. It is critical to clarify that the control outcomes from this experimental condition are unrealistic due to the lack of data transfer, but the runtime performance is sufficient to offer insight into the extent of performance gains achievable.

The third set of experiments (\textit{Optimized}) is directed towards achieving an optimized state of I/O operations in the overall framework. The optimization measures implemented include, but are not limited to, the removal of unnecessary I/O of flow field data to reduce the number of files, and the use of binary file formats to reduce file sizes. The crux of this approach is to pare down the I/O to its functional essence, ensuring that only the necessary data exchanges that are crucial for the simulation and learning process are preserved, thereby maintaining the integrity and accuracy of the DRL and CFD simulations. \reply{With the applied I/O opitimization, we are able to reduce the file size from 5.0 MB to 1.2 MB for each CFD simulation, which is a significant 76\% reduction in data transfer volume.}

\begin{table}[ht]
\centering
\caption{Statistics of parallel multi-environment training for different I/O strategies: \textit{Baseline}, \textit{I/O-Disabled}, and \textit{Optimized}.}
\vspace{-\baselineskip}
\begin{tabularx}{0.95\textwidth}{
  >{\centering\arraybackslash}p{0.11\linewidth}
  >{\centering\arraybackslash}p{0.11\linewidth}
  >{\centering\arraybackslash}p{0.11\linewidth}
  >{\centering\arraybackslash}p{0.11\linewidth}
  >{\centering\arraybackslash}p{0.15\linewidth}
  >{\centering\arraybackslash}p{0.15\linewidth}
  >{\centering\arraybackslash}p{0.15\linewidth}
}
\toprule
$N_\text{episodes}$ & $N_\text{envs}$ & $N_\text{ranks}$ & $N_\text{total CPUs}$ & $T_\text{baseline}$ (h) & $T_\text{I/O-disabled}$ (h) & $T_\text{optimized}$ (h) \\
\midrule
3000 & 1  & 1 & 1  & 225.2 & 193.1 (14\%) & 200.0 (11\%)  \\ 
3000 & 2  & 1 & 2  & 123.7 & 104.7 (15\%) & 103.8 (16\%) \\ 
3000 & 4  & 1 & 4  & 64.6 & 53.4 (17\%) & 52.1 (19\%) \\ 
3000 & 6  & 1 & 6  & 44.4 & 35.5 (20\%) & 35.7 (20\%) \\ 
3000 & 8  & 1 & 8  & 33.9 & 26.3 (22\%) & 26.7 (21\%) \\ 
3000 & 10 & 1 & 10 & 26.3 & 21.3 (19\%) & 21.5 (18\%) \\ 
3000 & 20 & 1 & 20 & 14.2 & 11.3 (20\%) & 11.3 (20\%) \\ 
3000 & 30 & 1 & 30 & 9.6 & 7.9 (18\%) & 8.3 (14\%) \\ 
3000 & 40 & 1 & 40 & 9.0 & 6.4 (29\%) & 6.3 (30\%) \\ 
3000 & 50 & 1 & 50 & 8.1 & 5.5 (32\%) & 5.3 (35\%) \\ 
3000 & 60 & 1 & 60 & 7.6 & 4.8 (37\%) & 4.8 (37\%) \\ 
\bottomrule
\end{tabularx}
\label{tab:comparison_table}
\end{table}

The corresponding results are shown in 
\cref{tab:comparison_table}, where $T_\text{baseline}$, $T_\text{I/O-disabled}$ and $T_\text{optimized}$  denote the total training time for the \textit{Baseline}, \textit{I/O-Disabled}, and \textit{Optimized} tests, respectively. The relative speedup of \textit{I/O-Disabled} and \textit{Optimized} with respect to \textit{Baseline} is computed and included in \cref{tab:comparison_table} as percentages. 
A comparative analysis reveals that the suspension of I/O operations yields a consistent reduction in training duration across various number of environments ($N_\text{envs}$), indicative of the significant cost attributable to I/O processes in the baseline scenario. The degree of speedup realized through disabling I/O increases from 14\% to 37\%, highlighting the I/O-related bottlenecks that become more pronounced as the 
parallelization scales.

By optimizing I/O operations, the training time ($T_\text{optimized}$) exhibits a reduction that closely mirrors the \textit{I/O-Disabled} state. Notably, the \textit{Optimized} configuration could reach the theoretical maximum speedup observed in most of the \textit{I/O-Disabled} experiments, which underscores the efficacy of the optimization techniques applied. We observed that I/O optimization appears to be particularly effective when $N_\text{envs}$ is very high. This is because the amount of I/O is directly related to the number of environments since each environment would require one set of files for data exchange with the DRL agent. With few environments, the amount of data movement is insufficient to saturate the disk I/O bandwidth and I/O costs are negligible. For large-scale parallelization, I/O bottlenecks emerge due to large amount of data transfer for multiple environments. The optimization techniques employed $--$ such as minimizing file numbers and transitioning to binary formats $--$ significantly alleviate the I/O overhead, thereby enhancing the overall computational efficiency.
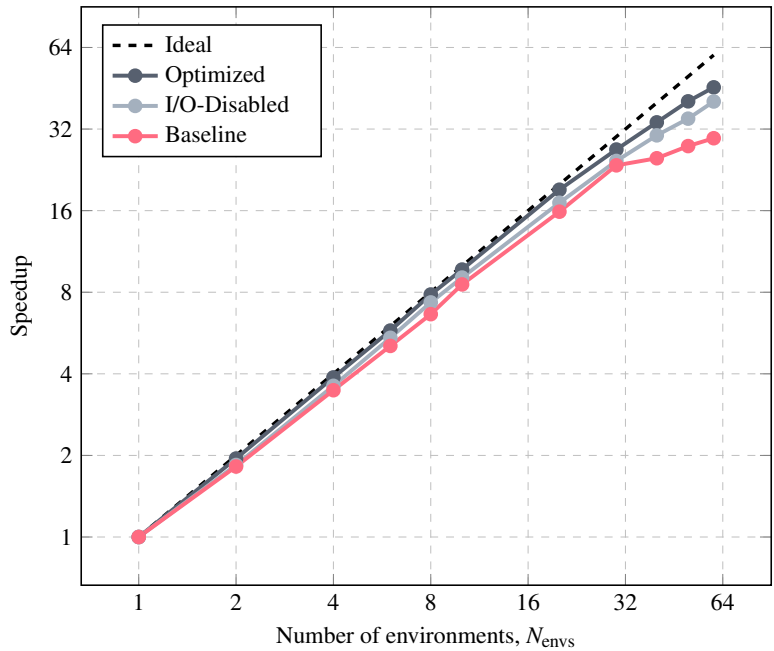
\begin{figure}[htbp]
\centering
\begin{tikzpicture}
\begin{loglogaxis}[
width=0.6\textwidth,
xtick={1,2,4,8,16,32,64},
xticklabels={1,2,4,8,16,32,64},
ytick={1,2,4,8,16,32,64},
yticklabels={1,2,4,8,16,32,64},
xlabel={Number of environments, $N_\text{envs}$},
ylabel={Speedup},
grid=both,
legend pos=north west,
legend style={nodes={text width=1.9cm}},
grid=major,
grid style={line width=0.25pt, dashed},
]
\addplot[black,dashed,line width=1.3pt] table [x=env, y=Perfect] {Figure/fig9/all_speedup.dat};
\addplot[hc22,mark=*,line width=1.5pt] table[x=Perfect, y=Optimized_08] {Figure/fig9/all_speedup.dat};
\addplot[hc21,mark=*,line width=1.5pt] table[x=Perfect, y=No_IO] {Figure/fig9/all_speedup.dat};
\addplot[hc20,mark=*,line width=1.5pt] table[x=Perfect, y=Baseline] {Figure/fig9/all_speedup.dat};
\legend{Ideal, Optimized, I/O-Disabled, Baseline}
\end{loglogaxis}
\end{tikzpicture}
\caption{Speedup ratio for three I/O strategies under single-core CFD ($N_\text{ranks}=1$) and multi-environment training: \textit{Baseline}, \textit{I/O-Disabled}, and \textit{Optimized}. $N_\text{envs}=1$ in each set of data is taken as the respective reference point.}
\label{fig:parallel speedup}
\end{figure}

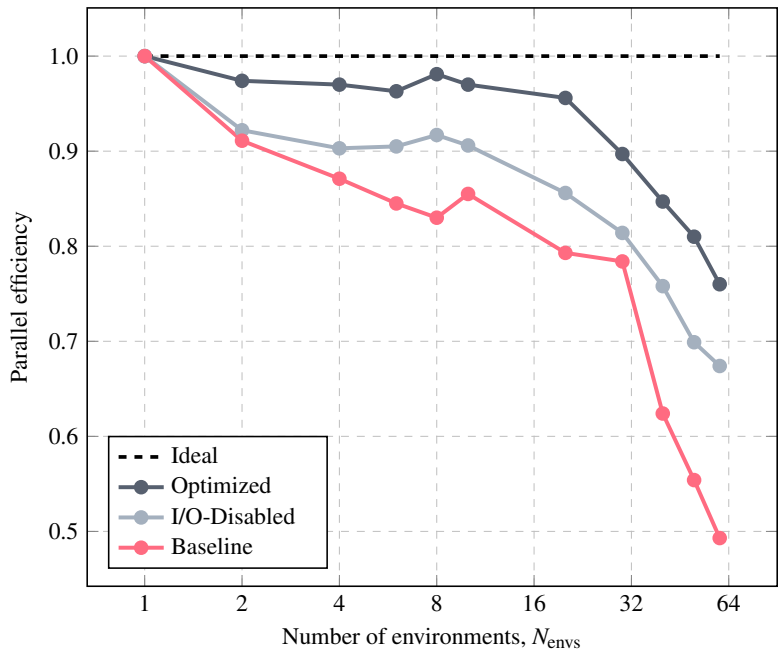
\begin{figure}[htbp]
\centering
\begin{tikzpicture}
\begin{semilogxaxis}[
width=0.6\textwidth,
xtick={1,2,4,8,16,32,64},
xticklabels={1,2,4,8,16,32,64},
ytick={0.5,0.6,0.7,0.8,0.9,1.0},
yticklabels={0.5,0.6,0.7,0.8,0.9,1.0},
xlabel={Number of environments, $N_\text{envs}$},
ylabel={Parallel efficiency},
grid=both,
legend pos=south west,
legend style={nodes={text width=1.9cm}},
grid=major,
grid style={line width=0.25pt, dashed},
]
\addplot[black, dashed, line width=1.4pt] table [x=env, y=Perfect] {Figure/fig10/all_paraller_efficiency.dat};
\addplot[hc22, mark=*,line width=1.5pt] table[x=env, y=Optimized_08] {Figure/fig10/all_paraller_efficiency.dat};  
\addplot[hc21, mark=*,line width=1.5pt] table[x=env, y=No_IO] {Figure/fig10/all_paraller_efficiency.dat};
\addplot[hc20, mark=*,line width=1.5pt] table[x=env, y=Baseline] {Figure/fig10/all_paraller_efficiency.dat};
\legend{Ideal, Optimized, I/O-Disabled,Baseline}
\end{semilogxaxis}
\end{tikzpicture}
\caption{Parallel efficiency for three I/O strategies under single-core CFD ($N_\text{ranks}=1$) and multi-environment training: \textit{Baseline}, \textit{I/O-Disabled}, and \textit{Optimized}. $N_\text{envs}=1$ in each set of data is taken as the respective reference point.}
\label{fig:parallel efficiency}
\end{figure}

The scaling performance of these experiments are visualized in~\cref{fig:parallel speedup} and \cref{fig:parallel efficiency}. The speedup ratio and parallel efficiency for each set of experiment is computed using the respective single-environment data of each set as the reference. In \cref{fig:parallel speedup}, the \textit{I/O-Disabled} strategy, which hypothetically eliminates all I/O constraints, closely tracks the perfect scaling line, particularly at lower environment counts. This close alignment diminishes slightly with a higher count of environments, suggesting that factors other than I/O, possibly computational overhead or inter-process communication, begin to influence performance at scale. The \textit{Optimized} curve showcases the practical application of I/O optimization techniques, which exhibits a substantial speedup that indicates the effectiveness of the optimizations in reducing I/O overhead. The \textit{Optimized} set of results appear above the \textit{I/O-Disabled} curve due to differences in the reference used for computing speedup. In \cref{fig:parallel efficiency}, it is evident that both strategies greatly improve the parallel efficiency. The \textit{Optimized} strategy is close to the ideal efficiency. Similar to the speedup computation, the parallel efficiency of \textit{Optimized} strategy is higher than the that of \textit{I/O-Disabled} because of the choice of reference. 
Nonetheless, the scaling trend and achievable parallel efficiency underscore the potential of I/O optimization in large-scale, parallel DRL and CFD computations to achieve near-ideal scaling efficiency.

In short, optimizing I/O processes within DRL-based CFD simulations is crucial as it significantly enhances computational efficiency, particularly in large-scale parallel computing environments where file I/O becomes a serious bottleneck. Such optimizations directly contribute to reducing training times and improving the scalability of simulations. In our tests, refining I/O operations improves the parallel efficiency significantly, from $\sim$49\% to $\sim$76\% on 60 cores. \reply{For larger problems with more significant disk footprint, the runtime performance regression associated with I/O operations will be more pronounced, since the bandwidth for hard disk is significant lower than that for cache or main memory. The imperative to implement I/O optimizations is markedly more critical for efficient scalability.}

\section{Conclusions}\label{sec:Conclusions} 

In this work, we have conducted a comprehensive investigation into the parallel scaling performance of DRL-based frameworks for AFC problems, shedding light on key factors influencing computational efficiency in high-performance computing environments. 
\begin{itemize}
    \item We validated a state-of-the-art DRL framework for a typical AFC problem of flow past a circular cylinder and showcased its capability to achieve drag reduction within a few hundreds of episodes. In addition, we verified that parallelization through multiple environments gives rise to similar reward convergence, indicating the potential of speedup through parallelization.
    \item We deconstructed the framework and analyzed parallel efficiency of the CFD component. The parallel efficiency of CFD drops rapidly in multi-core settings, reaching less than 20\% for 16 MPI ranks.
    \item On the other hand, the parallel efficiency of multi-environment training is much higher, where 60 single-core environments could reach 49\% efficiency. By extensive benchmarking of various hybrid parallelization configurations, we identified an optimal hybrid parallelization strategy that strongly favors parallelizing DRL across multiple environments over parallelizing CFD simulations. With this parallelization strategy, the training time can be reduced from 225.2 hours ($N_\text{ranks}=1, N_\text{envs}=1$) to 7.6 hours ($N_\text{ranks}=1, N_\text{envs}=60$), leading to a speedup of 30 times.
    \item To further optimize the parallelization performance, this study brought to the forefront the critical issue of I/O operations, a previously under-emphasized aspect in such simulations. By implementing targeted solutions to mitigate I/O costs, we observed a substantial enhancement in scaling efficiency from 49\% to 78\%. The final optimal training time is reduced to 4.8 hours ($N_\text{ranks}=1, N_\text{envs}=60$ with \textit{I/O Optimized}), which is equivalent to a remarkable speedup of approximately 47 times in runtime performance.
\end{itemize}
Overall, we were able to achieve near-linear scaling performance and significantly boost parallel efficiency, underscoring the effectiveness of the optimizations introduced.

It is important to acknowledge that there are still many unknowns and uncertainties when it comes to applying the frameworks and methodologies discussed here to different DRL-based AFC problems. Nonetheless, the theoretical trend uncovered here - that parallelizing DRL through multiple environments is generally advantageous due to reduced data dependencies - could potentially be extrapolated to other frameworks. This is particularly relevant unless the CFD component itself can be parallelized with very high efficiency comparable to that of multi-environment parallelism. Additionally, this research highlights the importance of not overlooking I/O issues, especially in situations where there is substantial data transfer between the DRL and CFD components. We believe these insights are significant for the field of DRL-based CFD simulations and provide a foundation for future directions, such as asynchronous reinforcement learning training in AFC problems.

\section*{Declaration of interests}
The authors report no conflict of interest.

\section*{References}
\bibliography{aipsamp}

\end{document}